%% file: main.tex
\documentclass[10pt,twocolumn,letterpaper]{article}

\usepackage{cvpr}              %

\usepackage{bm}
\usepackage{graphicx}
\usepackage{amsmath}
\usepackage{amssymb}
\usepackage{booktabs}
\usepackage{multirow}
\usepackage{tcolorbox}
\usepackage{color}
\usepackage{siunitx}
\usepackage{colortbl}
\usepackage{listings}
\usepackage{algorithmic}
\usepackage{algorithm}
\usepackage[font={small}]{caption}

\definecolor{royalblue}{RGB}{65,105,225}
\definecolor{better}{RGB}{0, 165, 156}
\definecolor{tab_blue}{RGB}{31, 119, 180}
\definecolor{tab_orange}{RGB}{255, 127, 14}
\definecolor{tab_green}{RGB}{44, 160, 44}
\definecolor{tab_red}{RGB}{214, 39, 40}

\usepackage[pagebackref,breaklinks=true,colorlinks,citecolor=royalblue,bookmarks=false]{hyperref}

\usepackage[capitalize]{cleveref}
\crefname{section}{Sec.}{Secs.}
\Crefname{section}{Section}{Sections}
\Crefname{table}{Table}{Tables}
\crefname{table}{Tab.}{Tabs.}

\def\cvprPaperID{10106} %
\def\confName{CVPR}
\def\confYear{2023}

\def\nlp{NLP}
\def\tb{ETR}
\def\ourmethod{ETR-NLP}  %
\def\etal{{\it et al. }}

\begin{document}

\title{Mitigating Task Interference in Multi-Task Learning via Explicit Task Routing with Non-Learnable Primitives}

\author{Chuntao Ding$^{1}$\thanks{\emph{Work done as a visiting scholar at Michigan State University.}}\quad\quad Zhichao Lu$^{2}$\thanks{\emph{Corresponding author}}\quad\quad Shangguang Wang$^{3}$\quad\quad Ran Cheng$^{4}$\quad\quad Vishnu N. Boddeti$^{5}$\\
{\normalsize $^{1}$ Beijing Jiaotong University\quad $^{2}$ Sun Yat-sen University\quad $^{3}$ Beijing University of Posts and Telecommunications}\\
{\normalsize $^{4}$ Southern University of Science and Technology\quad $^{5}$ Michigan State University} \\
{\tt\small chuntaoding@163.com\quad\{luzhichaocn, ranchengcn\}@gmail.com\quad sgwang@bupt.edu.cn\quad vishnu@msu.edu}
}

\maketitle

\input{0-abstract.tex}
\input{1-introduction.tex}
\input{2-related-work.tex}
\input{3-method.tex}
\input{4-experiment.tex}
\input{5-conclusion.tex}

\section*{Acknowledgements}
This work was supported by the National Natural Science Foundation of China (No. 62202039, 62106097, 62032003) and the National Key Research and Development Program of China (No. 2022ZD0118502).

\appendix
\input{6-appendix}

{\small
\bibliographystyle{ieee_fullname}
\bibliography{egbib}
}

\end{document}

%% file: 0-abstract.tex
\begin{abstract}
Multi-task learning (MTL) seeks to learn a single model to accomplish multiple tasks by leveraging shared information among the tasks. Existing MTL models, however, have been known to suffer from negative interference among tasks. Efforts to mitigate task interference have focused on either loss/gradient balancing or implicit parameter partitioning with partial overlaps among the tasks. In this paper, we propose \ourmethod{} to mitigate task interference through a synergistic combination of non-learnable primitives (\nlp{}s) and explicit task routing (\tb{}). Our key idea is to employ non-learnable primitives to extract a diverse set of task-agnostic features and recombine them into a shared branch common to all tasks and explicit task-specific branches reserved for each task. The non-learnable primitives and the explicit decoupling of learnable parameters into shared and task-specific ones afford the flexibility needed for minimizing task interference. We evaluate the efficacy of \ourmethod{} networks for both image-level classification and pixel-level dense prediction MTL problems. Experimental results indicate that \ourmethod{} significantly outperforms state-of-the-art baselines with fewer learnable parameters and similar FLOPs across all datasets. Code is available at this \href{https://github.com/zhichao-lu/etr-nlp-mtl}{URL}.
\end{abstract}

%% file: 1-introduction.tex
\section{Introduction}

Multi-task learning (MTL) is commonly employed to improve learning efficiency and performance of multiple tasks by using supervised signals from other related tasks \cite{Zhang@ASurvey, Ruder@An, Crawshaw@Multi}. These models have led to impressive results across numerous tasks. However, there is well-documented evidence~\cite{Kokkinos@UberNet, maninis2019attentive, Zhao@AModulation, vandenhende2021multi} that these models are suffering from \emph{task interference}~\cite{Zhao@AModulation}, thereby limiting multi-task networks (MTNs) from realizing their full potential.

\begin{figure}[t]
    \begin{subfigure}[b]{0.28\linewidth}
        \centering
        \includegraphics[width=\textwidth]{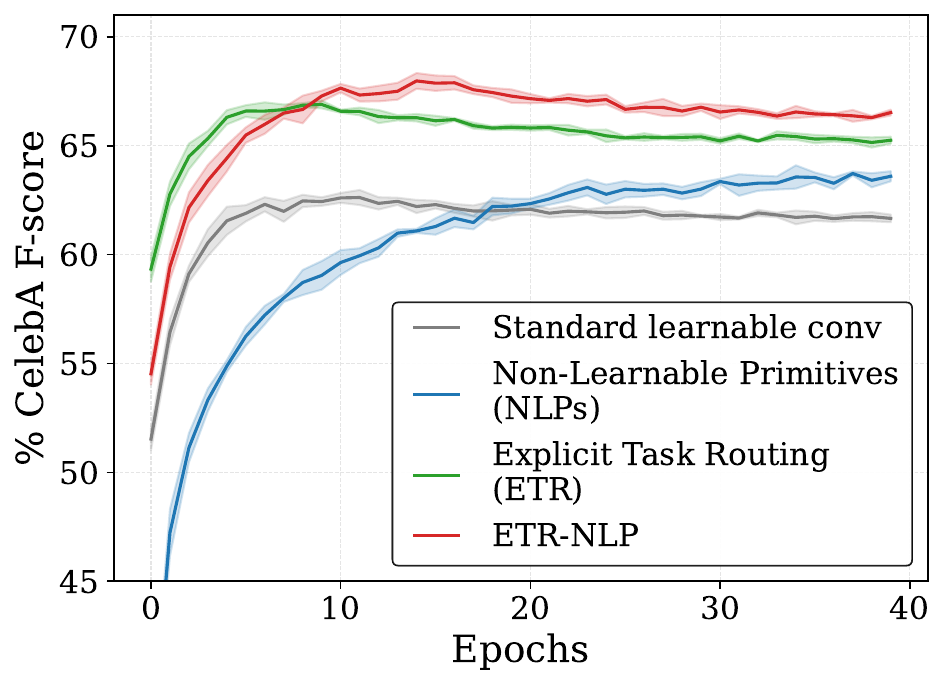}
        \vspace{-1.5em}
        \caption{\scriptsize ResNet18 \label{fig:convergence_conv_vs_nl_r18}}
    \end{subfigure}
    \begin{subfigure}[b]{0.73\linewidth}
        \centering
        \includegraphics[width=.32\textwidth]{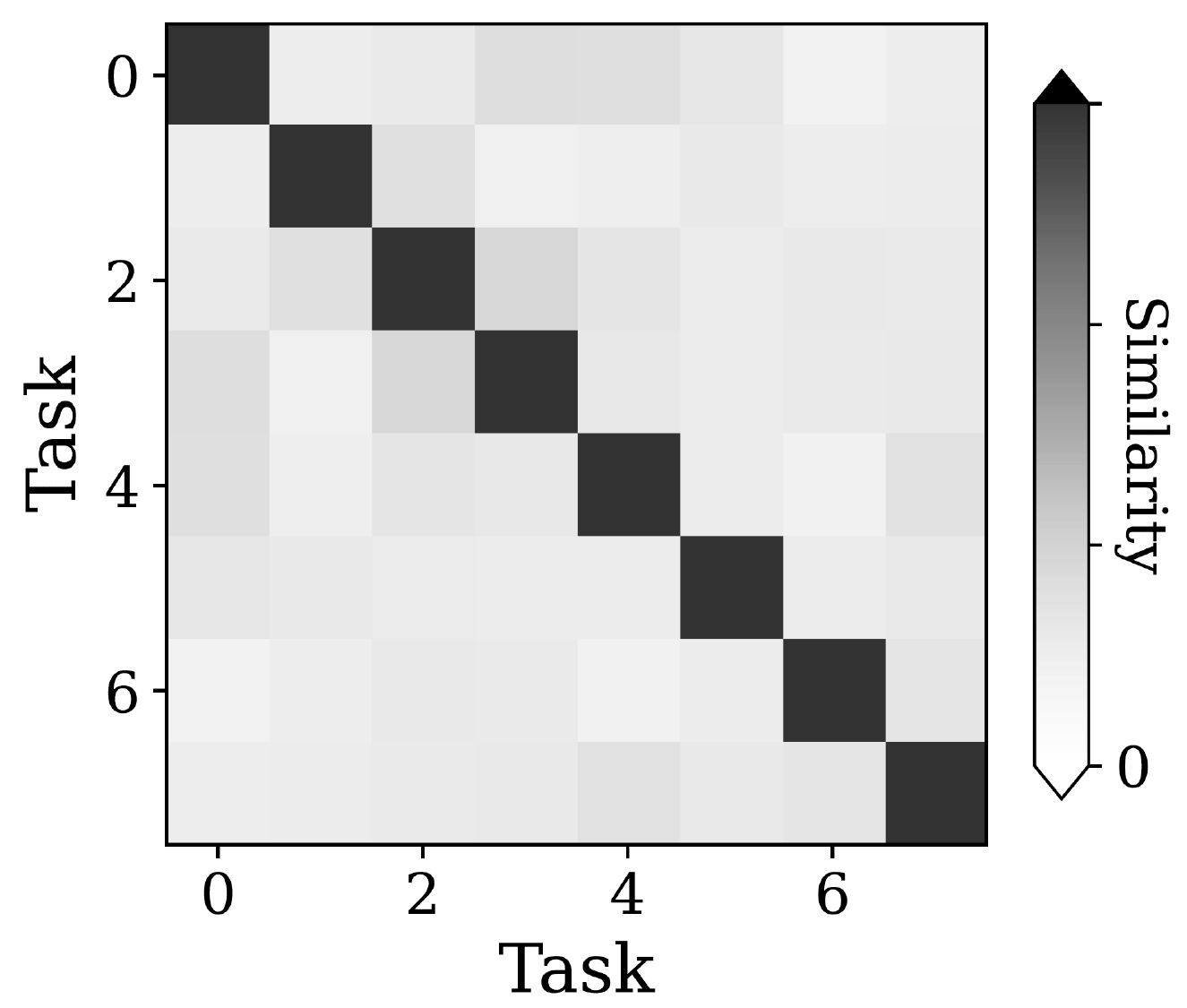}
        \includegraphics[width=.32\textwidth]{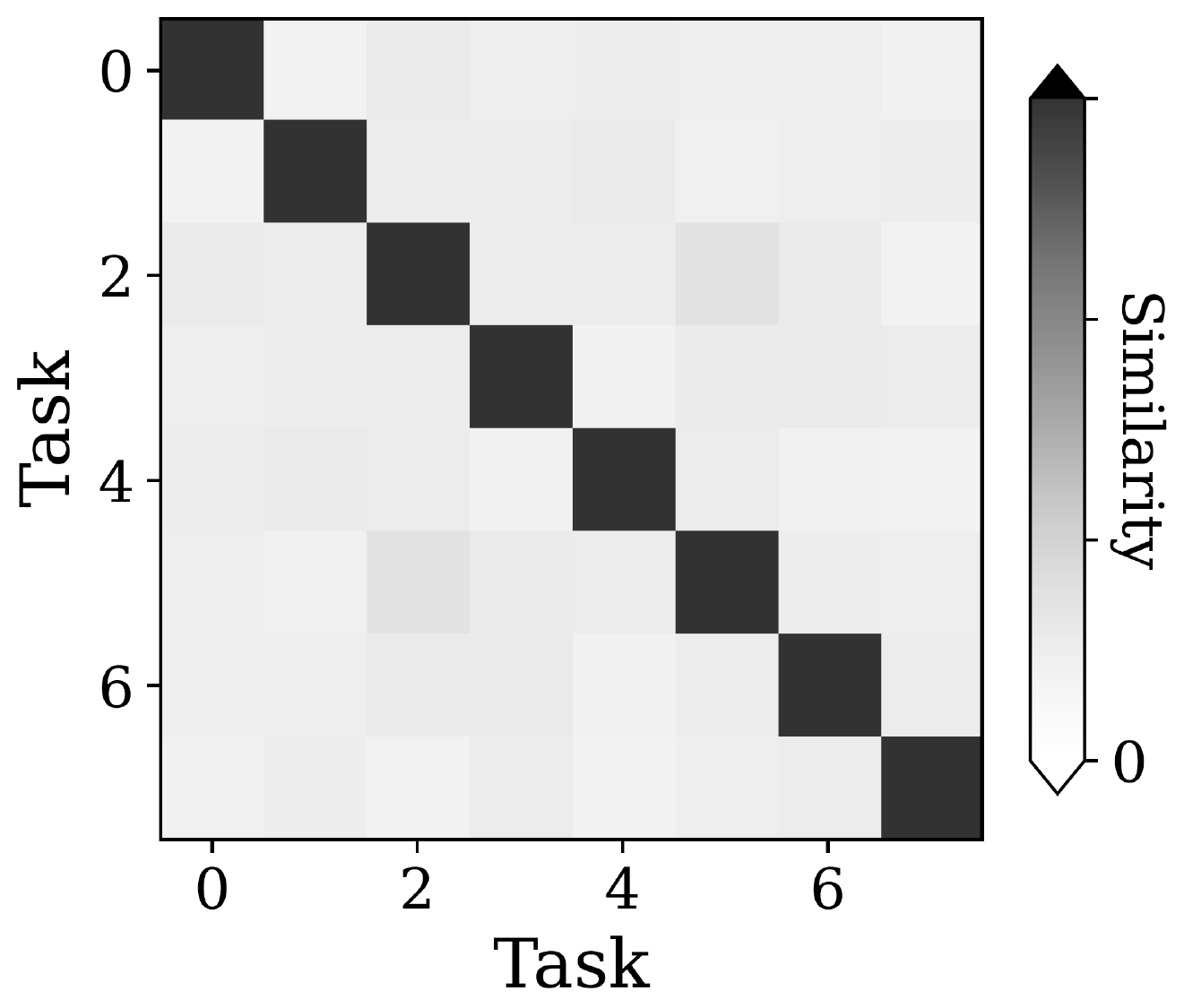}
        \includegraphics[width=.32\textwidth]{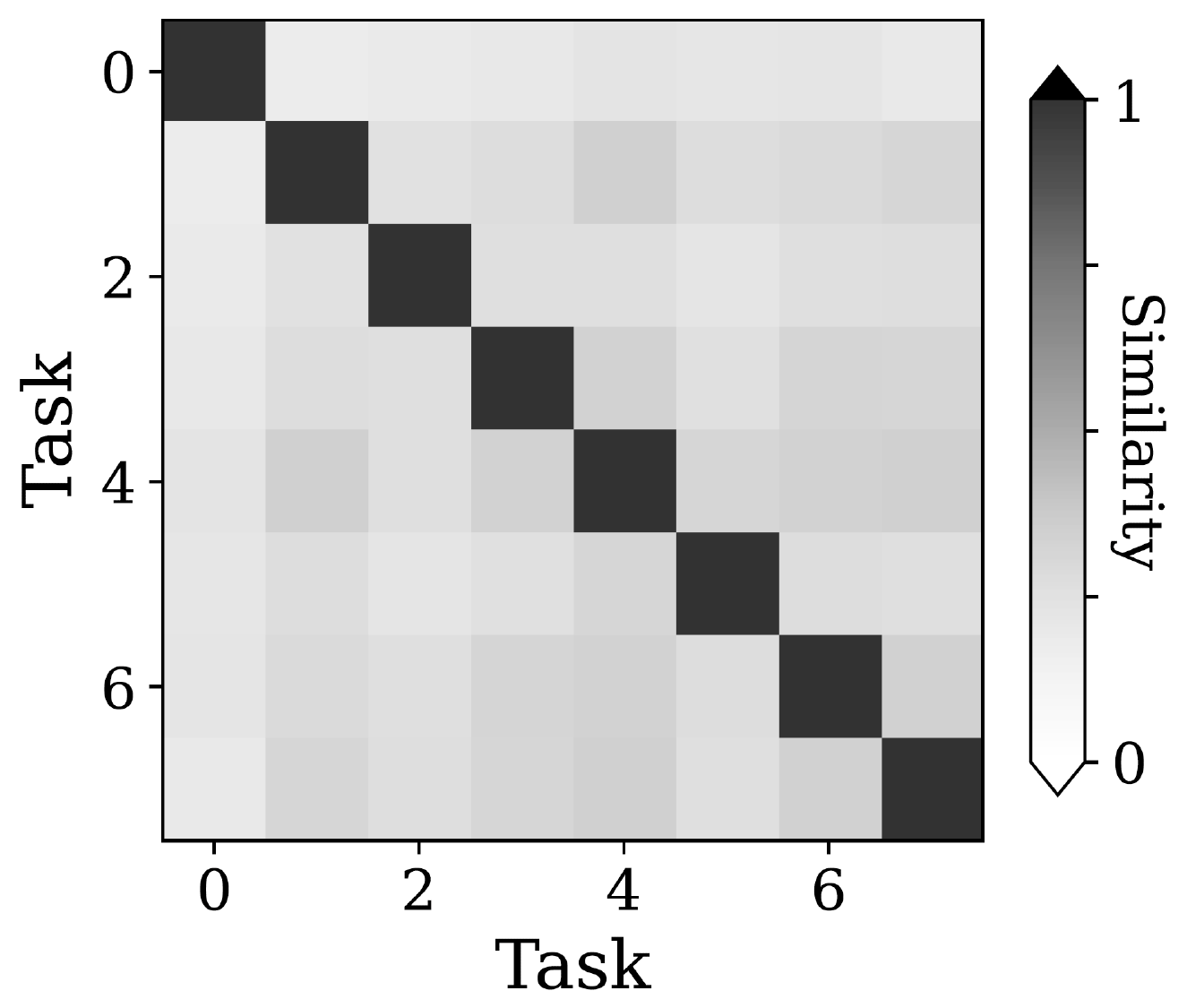} 
        \vspace{-.4em}
        \caption{\scriptsize Layer 1, 4, and 8 of ResNet18 (from left to right) \label{fig:celeba_r18_mtl_grad_cka}}
    \end{subfigure}
    \vspace{-2em}
    \caption{(a) Learning progression of multi-task networks (MTNs) on CelebA for eight tasks. Hard-sharing models with fully learnable parameters ({\color{gray}gray}) learn rapidly and then suffer from performance degradation due to conflicting gradients from task interference. Networks with non-learnable primitives (\nlp{}s; {\color{blue}{blue}}) do not suffer from task interference by design, while explicit task routing (ETR; {\color{green}green}), and ETR with NLPs ({\color{red}red}) do not eliminate but suffer less from task interference. (b) Gradient correlations measured via CKA~\cite{kornblith2019similarity} across all pairs of tasks for different layers of a standard MTN at the end of training. Observe the acute lack of correlation between tasks (low off-diagonal magnitude).\label{fig:teaser}\vspace{-0.5cm}}
\end{figure}
For instance, consider the learning progression of an MTN with a standard learnable convolutional layer in Figure~\ref{fig:convergence_conv_vs_nl_r18} ({\color{blue}blue curve}). Observe that the model learns rapidly, we posit, by exploiting all the shared information between the tasks, i.e., gradients pointing in similar directions. However, the performance starts degrading on further training since the model needs to exploit dissimilar information between the tasks for further improvement, i.e., gradients point in different directions. The latter can be verified by observing the similarity (centered kernel alignment~\cite{kornblith2019similarity}), or the lack thereof, between the gradients for each pair of tasks in Figure~\ref{fig:celeba_r18_mtl_grad_cka}.

Several approaches were proposed for mitigating task interference in MTNs, including loss/gradient balancing~\cite{Kendall@Multi,chen2018gradnorm,liu2019end, MGDA, Lin@Pareto}, parameter partitioning~\cite{Strezoski@Many, Bragman@Stochastic, maninis2019attentive,pascal2020maximum} and architectural design~\cite{Eigen@Predicting, Kokkinos@UberNet, Misra@Cross}. Despite the diversity of these approaches, they share two common characteristics, (i) all parameters are learned, either for a pre-trained task or for the multiple tasks at hand, (ii) the learned parameters are either fully shared across all tasks or are shared across a partial set of tasks through implicit partitioning, i.e., with no direct control over which parameters are shared across which tasks. Both of these features limit the flexibility of existing multi-task network designs from mitigating the deleterious effects of task interference on their predictive performance. 

\begin{figure}[t]
\centering
\includegraphics[width=.485\textwidth]{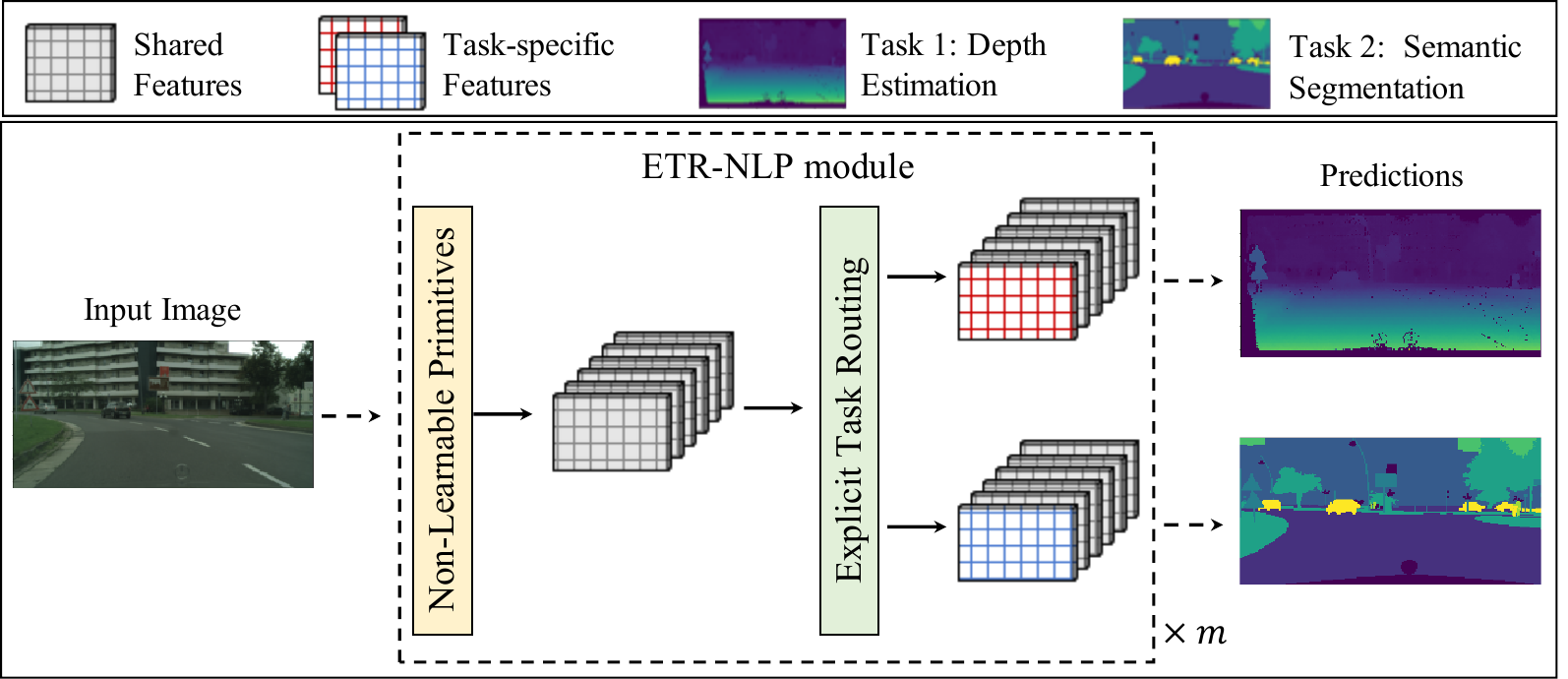}
\vspace{-2em}
\caption {\textbf{\ourmethod{} Networks} comprise non-learnable primitives to extract diverse task-agnostic features, followed by explicit task routing to control the parameters/features that are shared across all tasks and those that are exclusive to every single task.\label{fig:overview}\vspace{-0.5cm}}
\end{figure}

Relaxing the above design choices is the primary goal of this paper. We propose two complementary design principles, namely \emph{explicit task routing (\tb{})}, and \emph{non-learnable primitives (\nlp{}s)}, that explicitly seek to mitigate task interference. Through extensive empirical evaluation, we demonstrate that these two complementary ideas, individually and jointly, help mitigate task interference and consistently improve the performance of MTNs. As can be observed in Figure~\ref{fig:convergence_conv_vs_nl_r18}, compared to a hard-sharing MTN with a standard learnable convolutional layer ({\color{gray}gray curve}), an MTN with \nlp{} ({\color{blue}blue curve}) has better learning characteristics, i.e., learn more steadily and not suffer from performance degradation. Similarly, MTN with \tb{} ({\color{green}green curve}) and MTN with \ourmethod{} ({\color{red}red curve}) does not eliminate task interference but reduce it to an extent. Figure~\ref{fig:overview} shows an overview of the proposed \ourmethod{} networks.

From a network topological perspective, we propose explicit task-routing (\tb{}), a parameter allocation strategy that partitions the parameters into \emph{shared} and \emph{task-specific} branches. More explicitly, it comprises one branch shared across all tasks and task-specific branches, one for each task. Unlike existing parameter partitioning methods, \tb{} is designed to offer precise and fine-grained control over \emph{which} and \emph{how many} parameters are \emph{shared} or \emph{not shared} across the tasks. Additionally, \tb{} is flexible enough to allow existing implicit parameter partitioning methods \cite{Strezoski@Many,pascal2020maximum} to be incorporated into its shared branch.

From a layer design perspective, we propose using non-learnable primitives (\nlp{}s) to extract task-agnostic features and allow each task to choose optimal combinations of these features adaptively. There is growing evidence that features extracted from \nlp{}s can be very effective for single-task settings, including for image classification~\cite{Xu@Local, XU@Perturbative, ramanujan2020s, Wu@Shift, yu2022metaformer}, reinforcement learning~\cite{gaier2019weight} and modeling dynamical systems~\cite{maass2002real}. \nlp{}s are attractive for mitigating task interference in MTL. Since they do not contain learnable parameters, the task-agnostic features extracted from such layers alleviate the impact of conflicting gradients, thus implicitly addressing task interference. However, the utility and design of \nlp{}s for multi-task networks have not been explored. We summarize our key contributions below:

\vspace{2pt}
\noindent\textbf{--} We introduce the concept of \emph{non-learnable primitives (\nlp{}s)} and \emph{explicit task routing (\tb{})} to mitigate task interference in multi-task learning. We systematically study the effect of different design choices to determine the optimal design of \tb{} and \nlp{}.

\vspace{2pt}
\noindent\textbf{--} We demonstrate the effectiveness of \tb{} and \nlp{} through MTNs constructed with only \nlp{}s (MTN-\nlp{}s) and only \tb{} (MTN-\tb{}) for both image-level classification and pixel-level dense prediction tasks.

\vspace{2pt}
\noindent\textbf{--} We evaluate the effectiveness of \ourmethod{} networks across three different datasets and compare them against a wide range of baselines for both image-level classification and pixel-level dense prediction tasks. Results indicate that \ourmethod{} networks consistently improve performance by a significant amount.

%% file: 2-related-work.tex
\section{Related Work}
We briefly review prior work on non-learnable primitives and mitigating task interference for multi-task learning (MTL), from which our work draws inspiration. Due to space constraints, we refer the readers to the supplementary material for more discussion of related work. We also encourage readers to refer to multiple excellent reviews~\cite{caruana1997multitask, Zhang@ASurvey, vandenhende2021multi} for a comprehensive overview of MTL.

\vspace{3pt}
\noindent {\bf{Non-Learnable Primitives (\nlp{}s):}} The notion of \nlp{} for feature extraction was explored for single-task learning motivated either by scientific curiosity or in the quest for computational efficiency \cite{Xu@Local, XU@Perturbative, Wu@Shift, Chen@All,gaier2019weight}. Xu et al. used non-learnable sparse binary convolutional filters referred to as \emph{LBConv} \cite{Xu@Local}. Wu et al. proposed randomly adjusting the spatial alignments of data, referred to as \emph{shift} \cite{Wu@Shift}. Xu et al. applied non-learnable additive noises sampled from a uniform distribution to data, referred to as \emph{perturbation} \cite{XU@Perturbative}. Yu et al. replaced the attention-based token mixer with a non-parametric pooling primitive in vision transformers \cite{yu2022metaformer}. As a common practice, a follow-up $1\times1$ convolution was used to learn a weighted linear combination of features extracted by non-learnable primitives. These methods generally perform as well as those with standard learnable layers but with much fewer parameters required to optimize. However, the utility of \nlp{}s for MTL is yet to be explored.

The \nlp{}-based feature extraction proposed in this work is notably different in three key respects: (i) We expand the scope of \nlp{} from single-task image classification to MTL, including image-level and pixel-level prediction tasks. (ii) We consider multiple types of \nlp{}s (i.e., pooling, shift, perturbation, convolution without learnable weights) and demonstrate that a single type of \nlp{} does not benefit MTL. (iii) We design an MTL-specific \nlp{} by exploring various combinations of \nlp{}s under a diverse set of hyperparameters (e.g., pooling/kernel size, sparsity, perturbation strength, real, binary, depth-wise separable weights, etc.).

\vspace{2pt}
\noindent{\textbf{Task Interference in MTL:}} The success of MTL in computer vision has led to many solutions for mitigating task interference in MTL. The approaches fall into three main categories, (i) loss/gradient balancing~\cite{Kendall@Multi,chen2018gradnorm,liu2019end, MGDA, Lin@Pareto}, (ii) parameter partitioning \cite{Strezoski@Many, Bragman@Stochastic, maninis2019attentive,pascal2020maximum}, and (iii) architectural design \cite{Eigen@Predicting, Kokkinos@UberNet, Misra@Cross} (supplementary materials). A brief overview is provided below.

\vspace{2pt}
\noindent\textbf{--}\emph{Loss/Gradient balancing:} 
Kendall \etal \cite{Kendall@Multi} utilized homoscedastic uncertainty as task-dependent weights to balance the losses of various tasks. Chen \etal \cite{chen2018gradnorm} adaptively balanced the training of deep MTL models by dynamically adjusting the magnitudes of gradients computed w.r.t different tasks. Liu \etal \cite{liu2019end} proposed task-specific loss functions to maintain balance among tasks. Finally, Sener and Koltun \cite{MGDA} applied multi-objective optimization to find Pareto-optimal gradients for multiple tasks. The primary goal of this class of methods is to control the contribution of the loss/gradient of each task to the overall loss/gradient, which in turn implicitly helps mitigate task interference.

\vspace{2pt}
\noindent\textbf{--}\emph{Parameter partitioning:} Attention mechanisms have been widely used to allow networks to focus on different regions of the feature maps adaptively \cite{vaswani2017attention}. Attention mechanisms have been employed~\cite{liu2019end} for MTL at the filter level, allowing each task to select a subset of parameters (i.e., partition) at each layer. Maninis \etal \cite{maninis2019attentive} used a task-specific squeeze-and-excitation module (i.e., channel attention) for soft parameter partitioning. Strezoski \etal \cite{Strezoski@Many} introduced a task routing module as a hard parameter partitioning strategy to alleviate interference among tasks by randomly assigning a sub-network to each task. Once assigned, the parameter partitioning remained unchanged. Maximum roaming improved task routing by adaptively updating the parameter partition assignments during training \cite{pascal2020maximum}. In contrast to the overlapped parameter partitioning in the aforementioned work, our task routing strategy explicitly reserves a task-specific branch of parameters exclusive for each task, leading to more precise control over the partitioning of parameters among the tasks.

%% file: 3-method.tex
\section{\ourmethod{} Network Design}
We first introduce non-learnable primitives (\nlp{}s) based feature extraction and explicit task routing (\tb{}). Then, we integrate both into a single module, dubbed \ourmethod{}, that can be incorporated into modern MTL architectures (e.g., ResNets \cite{He@Deep}, VGGs \cite{Simonyan@Very}, SegNet \cite{segnet}, etc.) in a straightforward manner. Lastly, we describe the network's training and inference process with \ourmethod{} modules.

\subsection{\nlp{} based Feature Extraction}

\nlp{}s afford several attractive properties that render them well-suited for MTL. \emph{Primarily}, \nlp{}s do not have any learnable parameters. Hence, the extracted features are agnostic to any particular task, alleviating the impact of disparate gradients. As such, \nlp{}s implicitly mitigate task interference and lead to better predictive performance. A \emph{secondary benefit} is from a computational perspective. Our proposed \nlp{} design offers computation benefits in terms of fewer learned parameters or lower FLOPs. However, as we demonstrate in \S \ref{sec:nl_results}, obtaining parameter efficiency in MTL is challenging since directly employing existing efficiency-oriented convolutional layer designs (which work very well on standard problems) leads to performance loss on MTL.

\begin{figure}[t]
    \begin{subfigure}[b]{0.15\textwidth}
    \centering
    \includegraphics[width=0.9\textwidth]{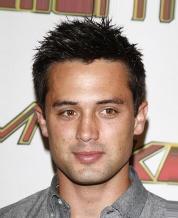}
    \caption{\scriptsize Inputs \label{fig:inputs}}
    \end{subfigure} \hfill
    \begin{subfigure}[b]{0.15\textwidth}
    \centering
    \includegraphics[width=0.9\textwidth]{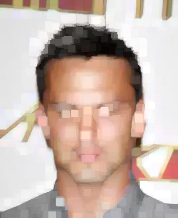}
    \caption{\scriptsize Max Pool \label{fig:nl_max}}
    \end{subfigure} \hfill
    \begin{subfigure}[b]{0.15\textwidth}
    \centering
    \includegraphics[width=0.9\textwidth]{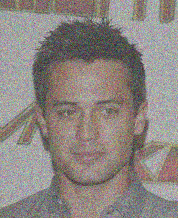}
    \caption{\scriptsize Perturbation \label{fig:nl_perturb}}
    \end{subfigure} \\
    \begin{subfigure}[b]{0.15\textwidth}
    \centering
    \includegraphics[width=0.9\textwidth]{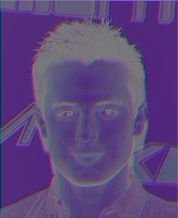}
    \caption{\scriptsize Conv (binary)\label{fig:nl_conv}}
    \end{subfigure} \hfill
    \begin{subfigure}[b]{0.15\textwidth}
    \centering
    \includegraphics[width=0.9\textwidth]{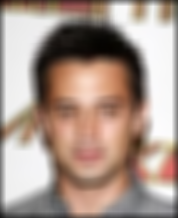}
    \caption{\scriptsize Avg. Pool \label{fig:nl_avg}}
    \end{subfigure} \hfill
    \begin{subfigure}[b]{0.15\textwidth}
    \centering
    \includegraphics[width=0.9\textwidth]{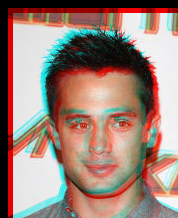}
    \caption{\scriptsize Shift\label{fig:nl_shift}}
    \end{subfigure}
    \vspace{-1em}
    \caption{\textbf{Visualization of features extracted by various \nlp{}s:} Given an input image (a), different types of \nlp{} extract diversely different features (b) - (f). \label{fig:nl_layer_visualization}\vspace{-0.5cm}}
\end{figure}

We hypothesize that extracting features from non-learnable primitives (\nlp{}s) that are neither biased nor adaptable to the tasks at hand can mitigate task interference in MTL. A plethora of \nlp{}s are available, including both non-parametric (e.g., average/max pooling, identity mapping, etc.) and weight-agnostic ones (e.g., LBConv \cite{Xu@Local}, perturbation \cite{XU@Perturbative}, shift~\cite{Wu@Shift, Chen@All}, etc.). Furthermore, one can tune each type of \nlp{} by adjusting its hyperparameters, such as pooling size, perturbation strength, the sparsity of the non-learnable weights, etc.

\begin{figure}[t]
    \centering
    \includegraphics[width=.45\textwidth]{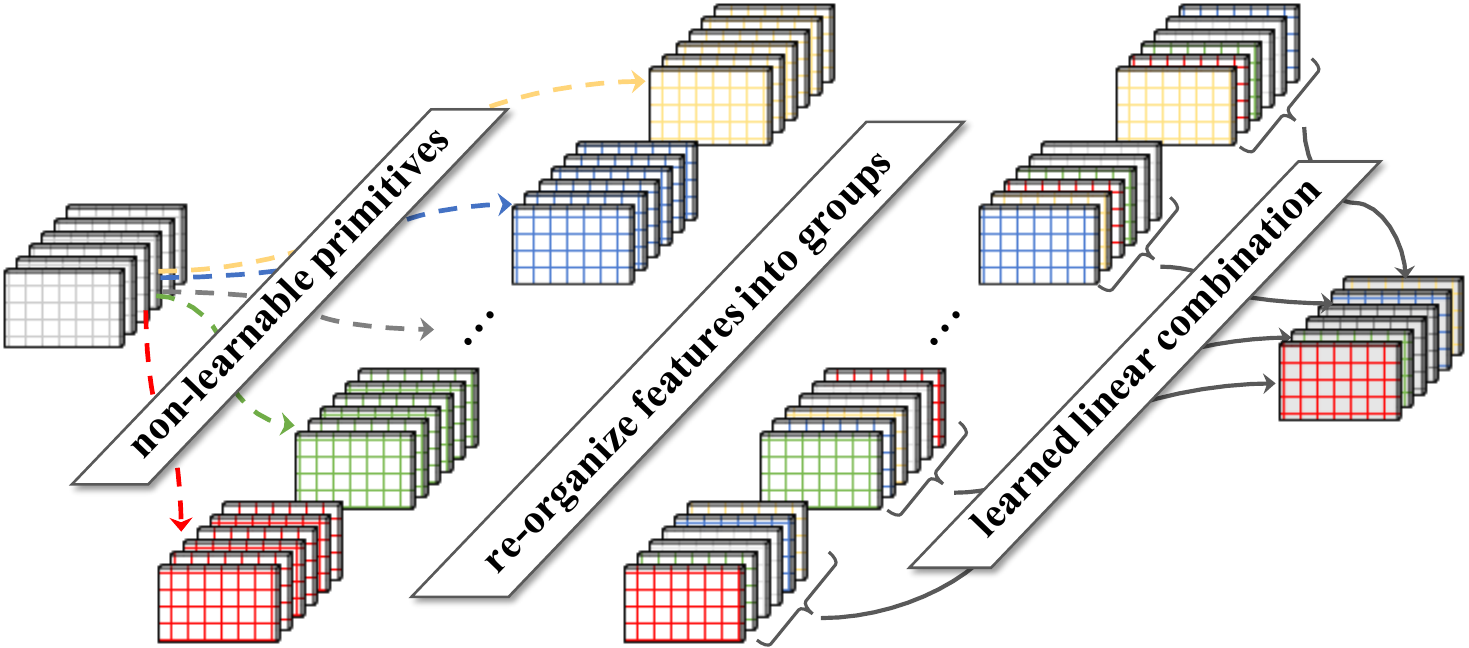}
    \vspace{-.6em}
    \caption{\textbf{Design of our \nlp{}-based feature extraction:} It first uses various types of non-learnable primitives (e.g., pooling, shift \cite{Wu@Shift, Chen@All}, perturbation \cite{XU@Perturbative}, etc.) with different hyperparameters (e.g., kernel size, sparsity, perturbation strength, etc.) to extract a dictionary of diverse features. Then, these features are re-arranged into groups by taking one output feature from each primitive. Subsequently, a linear combination is learned to compress features in a group into a single output feature. Finally, these compressed features from different groups are concatenated together. \label{fig:non_learnable_layer}} 
    \vspace{-1em}
\end{figure}

However, we demonstrate in Table~\ref{tab:abl_nl_combinations} directly employing a single type of non-learnable primitive degrades the performance of the corresponding MTL model. Since different tasks benefit from different kinds of features, a single \nlp{} is sub-optimal for MTL. And, as we observe in Figure~\ref{fig:nl_layer_visualization}, different \nlp{}s extract different features. Therefore, to facilitate extracting a dictionary of diverse features, we place various types of \nlp{}s across different hyperparameter combinations in parallel, similar to an Inception structure \cite{Szegedy@Going}. Next, we re-arrange the extracted features into groups to enhance diversity by taking one feature map from each primitive. Then, a linear combination among features within a group is learned via group-wise $1\times1$ convolutions. A pictorial illustration of this process is provided in Figure~\ref{fig:non_learnable_layer}. 

Based on the schema described above, we first determine the optimal combination of \nlp{}s for MTL. Specifically, we consider five types of \nlp{}s, i.e., average pooling, max pooling, convolution with fixed weights \cite{Xu@Local}, shift \cite{Wu@Shift}, and perturbation \cite{XU@Perturbative}, for a total of $\sum_{i=1}^{5}\binom{5}{i}=31$ possible variations. Then, we evaluate each variation on both CelebA multi-attribute classification \cite{liu@Deep} and Cityscapes dense prediction (semantic segmentation and depth estimation)~\cite{Cordts@The} MTL problems, and perform five repetitions to account for performance fluctuations. A representative subset of results is presented in Table~\ref{tab:abl_nl_combinations} (see supplementary material for full results). We observed that using more \nlp{}s for extracting features generally leads to better MTL performance. In particular, the combination of averaging pool, convolution with non-learnable weights, and perturbation in parallel emerges as the top choice, i.e., our final configuration of \nlp{}s. The effects of the hyperparameters of \nlp{}s are presented in \S \ref{sec:ablation}. 

\begin{table}[ht]
\centering
\caption{\textbf{Effect of different configurations of \nlp{}s:} Relative \textcolor{better}{improvements}/\textcolor{red}{degradation} over standard learnable convolution on CelebA multi-attribute classification are highlighted in color. \label{tab:abl_nl_combinations}}
\vspace{-.5em}
\resizebox{.485\textwidth}{!}{%
\begin{tabular}{@{\hspace{2mm}}cccccccc@{\hspace{2mm}}}
\toprule
\multirow{2}{*}{\#Types} & \multicolumn{5}{c}{Non-Learnable Primitives} & \multirow{2}{*}{\begin{tabular}[c]{@{}c@{}}CelebA \\ F-score ($\uparrow$)\end{tabular}} & \multirow{2}{*}{\begin{tabular}[c]{@{}c@{}} $\Delta_{\rm{p}}$ \\ ($\uparrow$)\end{tabular}} \\ \cmidrule(lr){2-6}
 & Avg. pool & Max pool & Conv & Shift & Perturb &  &  \\ \midrule
\multirow{2}{*}{1} &  &  & \checkmark &  &  & $61.1_{\pm 0.2}$ & \textcolor{red}{-$3.0\%$} \\
 &  & \checkmark &  &  &  & $61.3_{\pm 0.1}$ & \textcolor{red}{-$2.7\%$} \\ \midrule
\multirow{3}{*}{2} & \checkmark & \checkmark &  &  &  & $61.6_{\pm 0.1}$ & \textcolor{red}{-$2.3\%$} \\
 &  &  & \checkmark & \checkmark &  & $62.2_{\pm 0.2}$ & \textcolor{red}{-$1.3\%$} \\
 & \checkmark &  &  & \checkmark &  & $62.4_{\pm 0.1}$ & \textcolor{red}{-$1.0\%$} \\ \midrule
\multirow{3}{*}{3} & \checkmark &  & \checkmark & \checkmark &  & $62.4_{\pm 0.1}$ & \textcolor{red}{-$0.9\%$} \\
 &  & \checkmark &  & \checkmark & \checkmark & $64.5_{\pm 0.2}$ & \textcolor{better}{+$2.4\%$} \\
 & \checkmark &  & \checkmark &  & \checkmark & $66.3_{\pm 0.3}$ & \textcolor{better}{+$5.2\%$} \\ \midrule
\multirow{2}{*}{4} & \checkmark &  & \checkmark & \checkmark & \checkmark & $65.0_{\pm 0.1}$ & \textcolor{better}{+$3.2\%$} \\
 & \checkmark & \checkmark & \checkmark &  & \checkmark & $64.1_{\pm 0.1}$ & \textcolor{better}{+$1.8\%$} \\ \midrule
5 & \checkmark & \checkmark & \checkmark & \checkmark & \checkmark & $64.1_{\pm 0.2}$ & \textcolor{better}{+$1.7\%$} \\ \midrule
 & \multicolumn{5}{c}{Standard learnable convolution} & $63.0_{\pm 0.2}$ & $0.0\%$ \\ \bottomrule
\end{tabular}%
}
\end{table}

\subsection{Explicit Task Routing (\tb{})}
Despite the extracted features being agnostic to any particular task, a standalone application of \nlp{}s does not proactively address task interference. Therefore, to complement \nlp{}-based feature extraction, we present a novel parameter partitioning method, dubbed \emph{explicit task routing} (\tb{}), to provide precise and fine-grained control over the sharing of parameters among tasks. 

\begin{figure*}[t]
    \begin{subfigure}[b]{0.3\textwidth}
    \centering
    \includegraphics[width=0.95\textwidth]{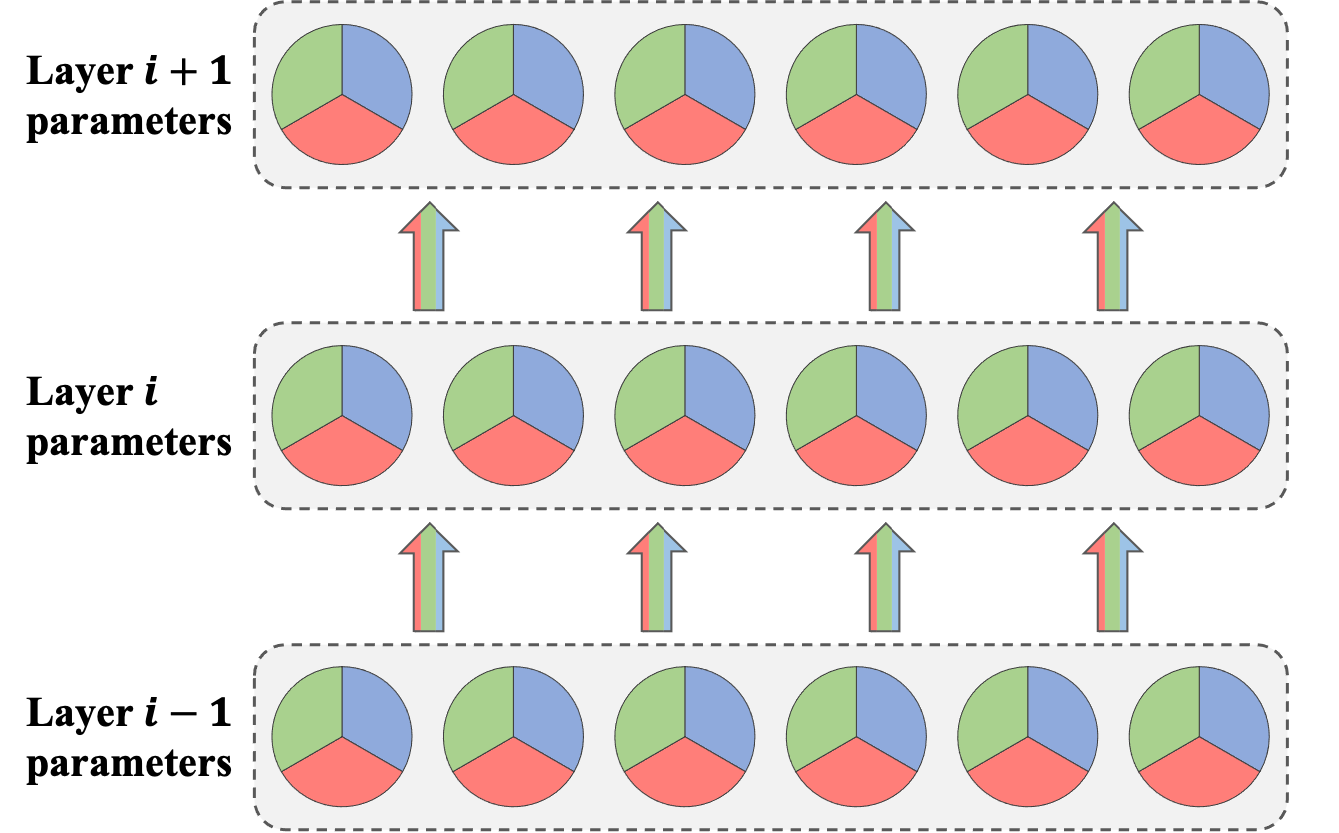}
    \caption{\scriptsize Hard parameter sharing \label{fig:mtl}}
    \end{subfigure} \hfill
    \begin{subfigure}[b]{0.3\textwidth}
    \centering
    \includegraphics[width=0.95\textwidth]{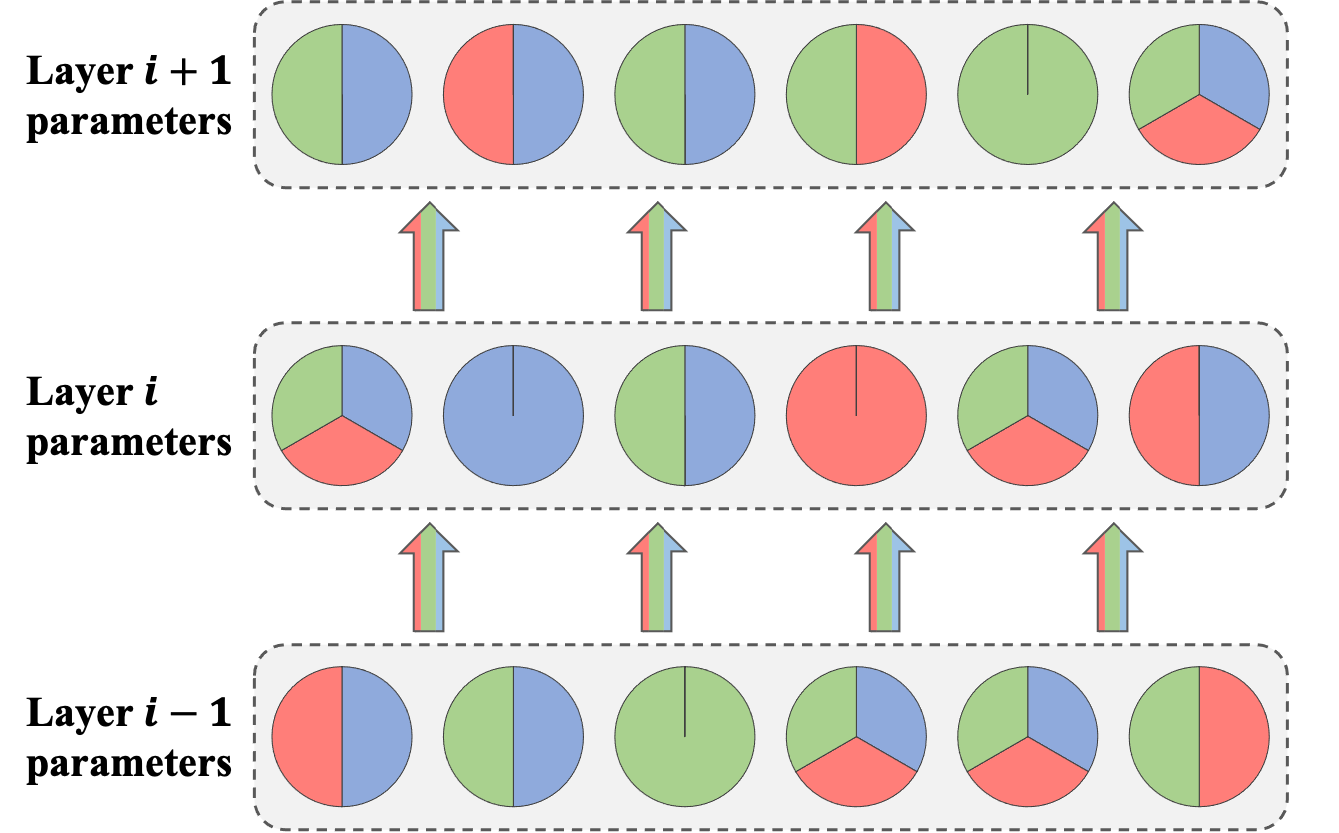}
    \caption{\scriptsize Existing parameter partitioning methods\label{fig:routing}}
    \end{subfigure} \hfill
    \begin{subfigure}[b]{0.375\textwidth}
    \centering
    \includegraphics[width=0.98\textwidth]{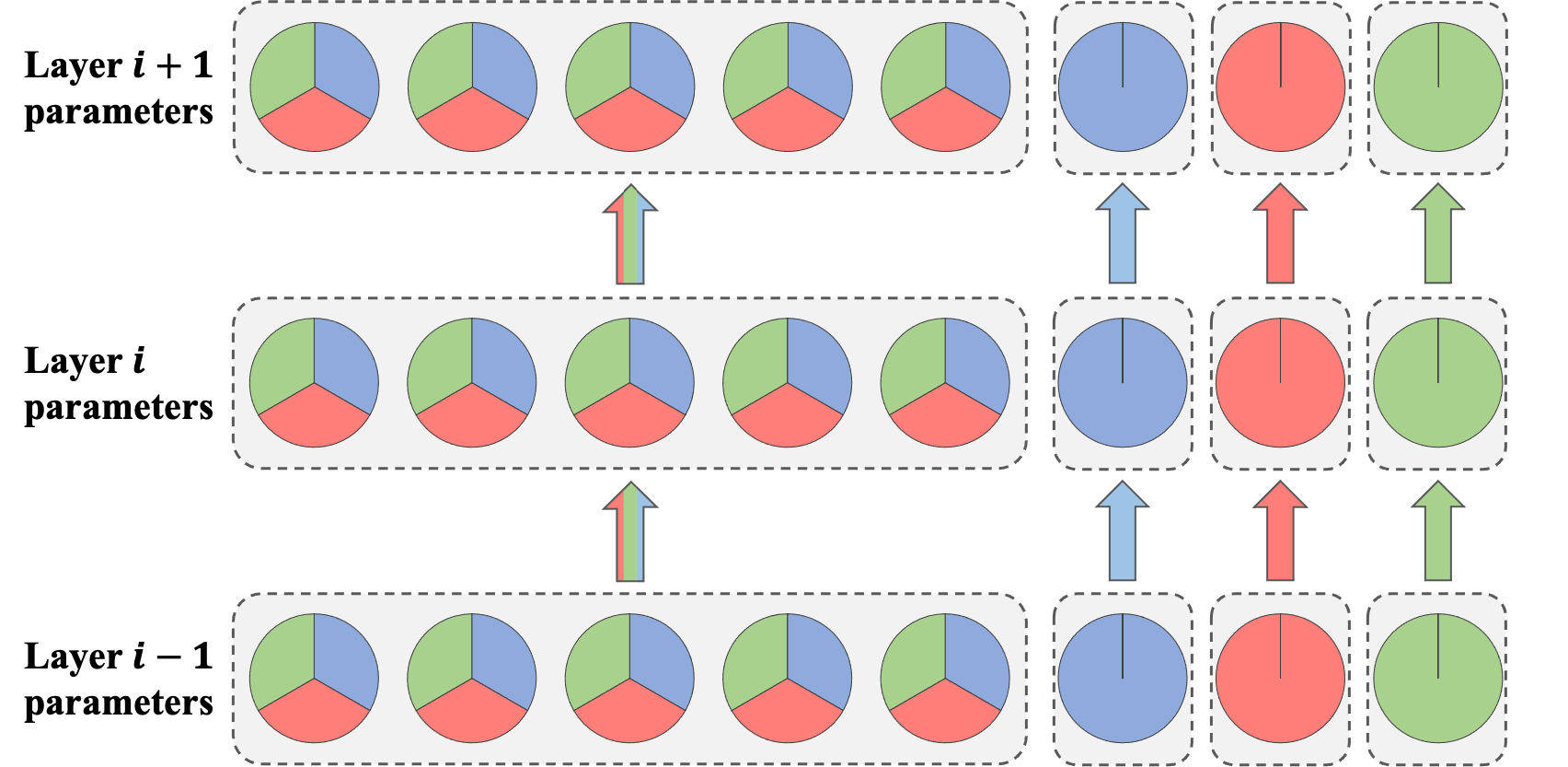}
    \caption{\scriptsize Parameter partitioning through explicit task routing (ours)\label{fig:branching}}
    \end{subfigure} 
    \vspace{-.5em}
    \caption{\textbf{Illustration of Explicit Task Routing (\tb{}):} (a) The de-facto hard parameter sharing method assigns all parameters to all tasks. (b) Existing parameter partitioning methods assign a partial and overlapped set of parameters for each task, and the assignment is either kept fixed \cite{Strezoski@Many} or updated/learned iteratively \cite{pascal2020maximum,maninis2019attentive}. (c) Our task explicit task routing separates parameters into a common branch shared by all tasks and a task-specific branch reserved exclusively for each task, providing more precise control over parameter partitioning among tasks. Note that only one task is activated during a forward pass for both the existing and our proposed parameter partitioning methods. \label{fig:mtl_vs_routing_vs_branching}\vspace{-0.2cm}}
\end{figure*}

Figure~\ref{fig:mtl_vs_routing_vs_branching} provides a pictorial illustration of \tb{} (along with the de-facto hard parameter sharing and existing parameter partitioning methods \cite{Strezoski@Many,pascal2020maximum}) for a three-task scenario. Parameters associated with the shared branch are shared across all tasks and absorb supervised signals common to all tasks. On the other hand, the parameters related to the task-specific branch are exclusive to each task and learn task-specific features. This explicit separation of parameters helps mitigate mutual interference among tasks. Additionally, to provide direct control and flexibility in terms of the number of shared parameters vs. task-specific parameters, we also introduce a hyperparameter, $\gamma \in [0,1]$, that controls the ratio of shared parameters over total available parameters. This ratio can be independently varied for each task. Note that $\gamma=0$ indicates the absence of the shared branch, which is equivalent to single-task learning. In contrast, $\gamma=1$ corresponds to all features shared among tasks, equal to a standard hard parameter sharing MTL. The effect of $\gamma$ is studied in \S \ref{sec:result_tab}.

\subsection{\ourmethod{} Module}
We incorporate the proposed \nlp{}s-based feature extraction and \tb{} parameter partitioning into a single module, dubbed \ourmethod{}, and form MTL networks by replacing the standard convolutional layers in modern MTL architectures with \ourmethod{} modules. In each \ourmethod{} module, since interactions between the tasks happen through the shared branch, the shared features are obtained by recombining task agnostic features through learned via group-wise $1 \times 1$ convolutions. Furthermore, since the task-specific weights are exclusive to each task and do not interfere with other tasks, the task-specific features are obtained by directly applying standard $3\times 3$ convolutions to features from the previous layer, i.e., task-specific features are not extracted for the task-specific branch. A pictorial illustration of \ourmethod{} and its corresponding pseudocode is shown in Figure~\ref{fig:teaser} and Algorithm~\ref{alg:code}, respectively.
\begin{algorithm}[t]
\caption{\ourmethod{}: PyTorch-like Pseudocode\label{alg:code}}
\definecolor{codeblue}{rgb}{0.25,0.5,0.5}
\definecolor{codekw}{rgb}{0.85, 0.18, 0.50}
\begin{lstlisting}[language=python, mathescape]
# C_in, C_out: number of input/output channels
# gamma: sharing ratio of explicit task routing
# prims: settings of non-learnable primitives
# T: No. of tasks
class NLP(nn.Module):
    def __init__(self, C_in, C_out, prims, **kwargs):
        # define non-learnable primitives
        for i, op in enumerate(prims):
            self.add_module(str(i), op(C_in, **kwargs))
        k = len(prims)  # No. of NLPs
        # group-wise linear combination 
        conv1x1 = nn.Conv2d(C_in * k, C_out, ks=1, groups=C_in)
        
    def forward(self, x):
        # extract features by NLPs
        y = torch.cat([op(x) for i, op in enumerate(self.values())], dim=1)
        # re-arrange features via channel shuffling 
        y = torch.channel_shuffle(y, groups=k)
        return conv1x1(y)  
        
class ETR-NLP(nn.Module):
    def __init__(self, C_in, C_out, gamma, prims, T, **kwargs):
        C_shared = int(gamma * C_out)
        C_specif = int(C_out - self.C_shared)
        # define a shared branch
        shared_branch = NLP(C_in, C_shared, prims, **kwargs)
        # define task-specific branches
        for i in range(T):
            specif_branch = nn.Conv2d(C_in, C_specif, ks=3, s=1, p=1)  # standard Conv
            self.add_module(``task_{}''.format(i), specif_branch)
        self.task = 0 # set an active task

    def get_layer(self, name):
        return getattr(self, name)
        
    def forward(self, x):
        shared = shared_branch(x)
        specif = self.get_layer(``task_{}''.format(self.task))(x)
        return torch.cat([shared, specif], dim=1)
\end{lstlisting}
\end{algorithm}

\noindent\textbf{{Training and inference:}} 
Similar to prior parameter partitioning-based methods \cite{maninis2019attentive, Strezoski@Many,pascal2020maximum}, only one task is activated at a time during a forward pass of our \ourmethod{}-based MTL networks. The training process for \ourmethod{}-based networks proceeds as follows. The shared branch and one task-specific branch are activated during a forward pass. As shown in Figure~\ref{fig:branching}, when task $i$ ($i \in [1,\mathcal{T}]$) is active, features for the $i\text{-th}$ task will be extracted through the shared and the active $i\text{-th}$ task-specific parameters. After training the current task, the parameters of the shared branch are updated immediately for image-level classification MTL problems (e.g., CelebA). While for dense prediction MTL problems (e.g., Cityscapes, NYU-v2), we wait until all tasks are forwarded before updating the parameters of the shared branch. These decisions are driven by an ablative analysis shown in \S \ref{sec:ablation}. During inference, a separate per-task evaluation is required as the input propagates through the shared and task-specific branches.

%% file: 4-experiment.tex
\section{Experimental Evaluation\label{ref:experiment}}
In this section, we first describe our experimental setup. Then, we independently demonstrate the effectiveness of \nlp{}s and \tb{} for MTL. Finally, we compare our \ourmethod{} to a wide range of MTL baselines for both image-level classification and pixel-level prediction problems.

\subsection{Experimental Setup\label{sec:setup}}

\noindent\textbf{Datasets.} We conduct experiments on three widely used MTL benchmarks: \emph{CelebA}~\cite{liu@Deep} is a large-scale face attributes dataset containing more than 200K celebrity images, each with 40 binary attribute annotations that can be grouped into eight categories. Accordingly, we can define an eight or 40-task MTL problem by considering each group or attribute as an individual binary classification task. \emph{Cityscapes}~\cite{Cordts@The} is a large-scale dataset for the semantic understanding of urban street scenes. It is split into training, validation, and test sets, with 2975, 500, and 1525 images. Following \cite{liu2019end,pascal2020maximum}, we resize all images to 128 by 256 and use the median level segmentation comprising seven semantic categories. Together with depth estimation, we define an eight-task MTL problem by treating the segmentation of each semantic category separately. \emph{NYU-v2} \cite{nyuv2} is a video sequence dataset composed of 1449 indoor images recorded over 464 scenes from a Microsoft Kinect camera. Following~\cite{liu2019end}, we resize all images to 288 by 384 resolution. It supports the segmentation of 13 semantic categories, depth estimation, and surface normal estimation for 15 tasks. More details are available in the supplementary.

\vspace{2pt}
\noindent\textbf{Implementation Details.} We implement \ourmethod{} in ResNet18 \cite{He@Deep} and VGG16 \cite{Simonyan@Very} architectures for image-level classification problems (e.g., CelebA), and in SegNet \cite{segnet} architecture for pixel-level dense prediction problems (e.g., Cityscapes and NYU-v2). For training on CelebA, we use Adam optimizer with a learning rate of $10^{-4}$ and a batch size of 256 images for 40 epochs. For training on Cityscapes and NYU-v2, we also use Adam optimizer with a learning rate of $10^{-4}$, but with a batch size of 8 images for 500 epochs. We repeat each experiment five times.

\vspace{2pt}
\noindent\textbf{Evaluation Metrics.} To evaluate the performance on image-level classification MTL problems, we consider \emph{precision}, \emph{recall}, and \emph{F-Score}. \emph{Precision} measures how precise a method is regarding how many predicted true instances are true positives. \emph{Recall} estimates how well a method has adapted to each task by measuring how much of the actual positive instances are recognized. \emph{F-Score} provides a composite measurement derived from precision and recall. This work reports mean precision, recall, and F-Score averaged over all tasks. For evaluating the performance on pixel-level dense prediction MTL problems, we track the \emph{mean Intersection over Union} (mIoU), and \emph{pixel accuracy} (Pix. Acc.) averaged over all segmentation tasks, and the \emph{mean absolute} (Abs. Err.) and \emph{relative error} (Rel. Err.) for the depth estimation task. Lastly, following prior work~\cite{caruana1997multitask, Zhang@ASurvey, vandenhende2021multi, Lin@Reasonable}, we also report the average relative improvement $\Delta_{\rm{p}}$ (defined below) w.r.t. a chosen baseline.

\footnotesize
$$ \Delta_{\rm{p}} = 100\% \times \frac{1}{T}\sum_{t=1}^{T}\frac{1}{N_t}\sum_{n=1}^{N_t} \frac{(-1)^{p_{t,n}}(M_{t,n} - M_{t,n}^{\rm{baseline}})}{M_{t,n}^{\rm{baseline}}}$$
\normalsize
\noindent where ${N_t}$ is the number of metrics in task $t$, $M_{t,n}$ is the performance of a task balancing method for the $n$-th metric in task $t$, $M_{t,n}^{\rm{baseline}}$ is defined similarly for the baseline method, and $p_{t,n}$ is one if a higher value indicates better performance for the $n$-th metric in task $t$ and zero otherwise.

\subsection{Experimental Results}

\subsubsection{Effectiveness of Non-Learnable Primitives \label{sec:nl_results}}
Table~\ref{tab:abl_nl_layer} compares our proposed NLPs with other alternative operations. We make the following observations, (i) All standalone instantiations of NLPs lead to performance degradation. The lack of diversity in the features is detrimental to predictive performance. (ii) Unlike standalone NLPs, our proposed NLP-based networks achieve higher precision and recall while requiring fewer learnable parameters and FLOPs. Since NLPs extract task-agnostic and diverse features, they can prevent the network parameters from being dominated by one or more tasks and mitigate mutual interference between tasks. Compared to the baseline architecture with standard learned convolution, NLPs-based networks significantly improve performance.
\begin{table}[t]
\centering
\caption{Comparison of NLPs with alternative designs on CelebA image-level classification problems. $2\times$ means a width multiplier of 2. Our results are highlighted with shading.\label{tab:abl_nl_layer}}
\vspace{-.5em}
\resizebox{.48\textwidth}{!}{%
\begin{tabular}{@{}llccccr@{\hspace{2mm}}}
\toprule
 & Method & $^{\#}$P & $^{\#}$F & Prec. ($\uparrow$) & Recall ($\uparrow$) & $\Delta_{\rm{p}}$ ($\uparrow$) \\ \midrule
\parbox[t]{2mm}{\multirow{7}{*}{\rotatebox[origin=c]{90}{ResNet18}}} & Conv & 11.2M & 148M & $67.7_{\pm{0.8}}$ & $59.8_{\pm{0.3}}$ & $0.0\%$ \\
 & LBConv \cite{Xu@Local} & 1.61M & 165M & $65.1_{\pm{0.8}}$ & $53.2_{\pm{0.4}}$ & \textcolor{red}{-$7.4\%$} \\
 & Shift \cite{Chen@All} & 2.81M & 42M & $67.5_{\pm{1.1}}$ & $58.4_{\pm{0.6}}$ & \textcolor{red}{-$1.3\%$} \\
 & Depth-wise Conv \cite{Howard@MobileNets} & 2.91M & 45M & $65.7_{\pm{0.4}}$ & $51.5_{\pm{0.4}}$ & \textcolor{red}{-$8.4\%$} \\
 & Ghost module \cite{han2020ghostnet} & 5.77M & 46M & $67.6_{\pm{1.5}}$ & $57.9_{\pm{0.6}}$ & \textcolor{red}{-$1.7\%$} \\
 & \cellcolor{gray!20}NLPs & \cellcolor{gray!20}2.05M & \cellcolor{gray!20}43M & \cellcolor{gray!20}$\bm{72.8_{\pm{0.3}}}$ & \cellcolor{gray!20}$59.2_{\pm{0.4}}$ & \cellcolor{gray!20}\textbf{\textcolor{better}{+$\bm{3.3\%}$}} \\
 & \cellcolor{gray!20}NLPs (2$\times$) & \cellcolor{gray!20}8.11M & \cellcolor{gray!20}148M & \cellcolor{gray!20}$71.3_{\pm{0.6}}$ & \cellcolor{gray!20}$\bm{62.2_{\pm{0.1}}}$ & \cellcolor{gray!20}\textbf{\textcolor{better}{+$\bm{4.7\%}$}} \\ \midrule
\parbox[t]{2mm}{\multirow{7}{*}{\rotatebox[origin=c]{90}{VGG16}}} & Conv & 14.7M & 1.25G & $71.1_{\pm{0.8}}$ & $63.8_{\pm{0.6}}$ & $0.0\%$ \\
 & LBConv \cite{Xu@Local} & 1.84M & 1.43G & $67.1_{\pm{0.6}}$ & $58.0_{\pm{0.4}}$ & \textcolor{red}{-$7.4\%$} \\
 & Shift \cite{Chen@All} & 1.67M & 0.14G & $68.9_{\pm{0.2}}$ & $59.5_{\pm{0.2}}$ & \textcolor{red}{-$4.9\%$} \\
 & Depth-wise Conv \cite{Howard@MobileNets} & 3.57M & 0.34G & $65.8_{\pm{0.7}}$ & $51.6_{\pm{0.8}}$ & \textcolor{red}{-$13.3\%$} \\
 & Ghost module \cite{han2020ghostnet} & 7.45M & 0.32G & $68.8_{\pm{0.8}}$ & $61.6_{\pm{0.4}}$ & \textcolor{red}{-$3.3\%$} \\
 & \cellcolor{gray!20}NLPs & \cellcolor{gray!20}2.48M & \cellcolor{gray!20}0.22G & \cellcolor{gray!20}$\bm{74.2_{\pm{0.8}}}$ & \cellcolor{gray!20}$64.5_{\pm{1.1}}$ & \cellcolor{gray!20}\textbf{\textcolor{better}{+$\bm{2.7\%}$}} \\
 & \cellcolor{gray!20}NLPs (2$\times$) & \cellcolor{gray!20}14.3M & \cellcolor{gray!20}1.23G & \cellcolor{gray!20}$72.5_{\pm{0.7}}$ & \cellcolor{gray!20}$\bm{68.7_{\pm{0.9}}}$ & \cellcolor{gray!20}\textbf{\textcolor{better}{+$\bm{4.8\%}$}} \\
 \bottomrule
\end{tabular}%
}
\vspace{-1em}
\end{table}

\subsubsection{Effectiveness of Explicit Task Routing \label{sec:result_tab}}
\begin{figure}[!ht]
    \begin{subfigure}[b]{0.485\textwidth}
    \centering
    \includegraphics[width=0.495\textwidth]{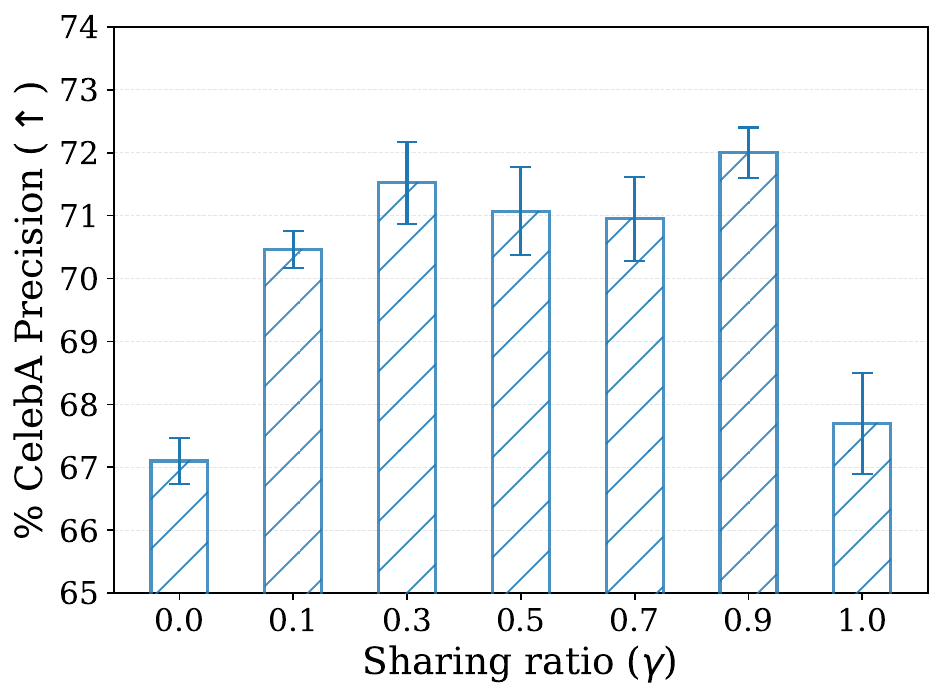}
    \hfill
    \includegraphics[width=0.495\textwidth]{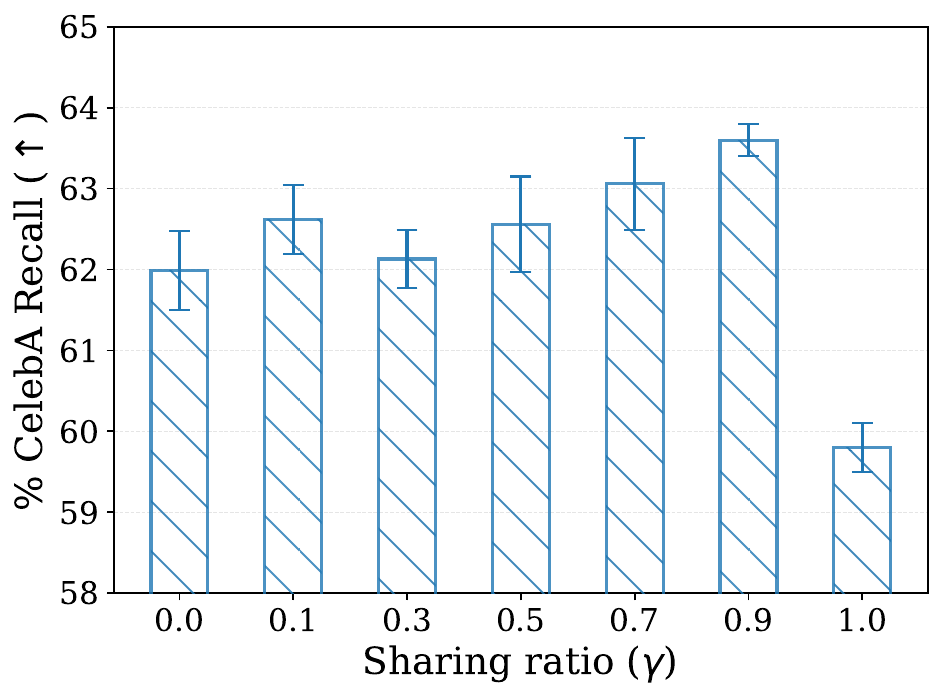}
    \caption{\scriptsize CelebA Image-Level Classification\label{fig:celeba_gamma}}
    \end{subfigure} \\
    \begin{subfigure}[b]{0.485\textwidth}
    \centering
    \includegraphics[width=0.495\textwidth]{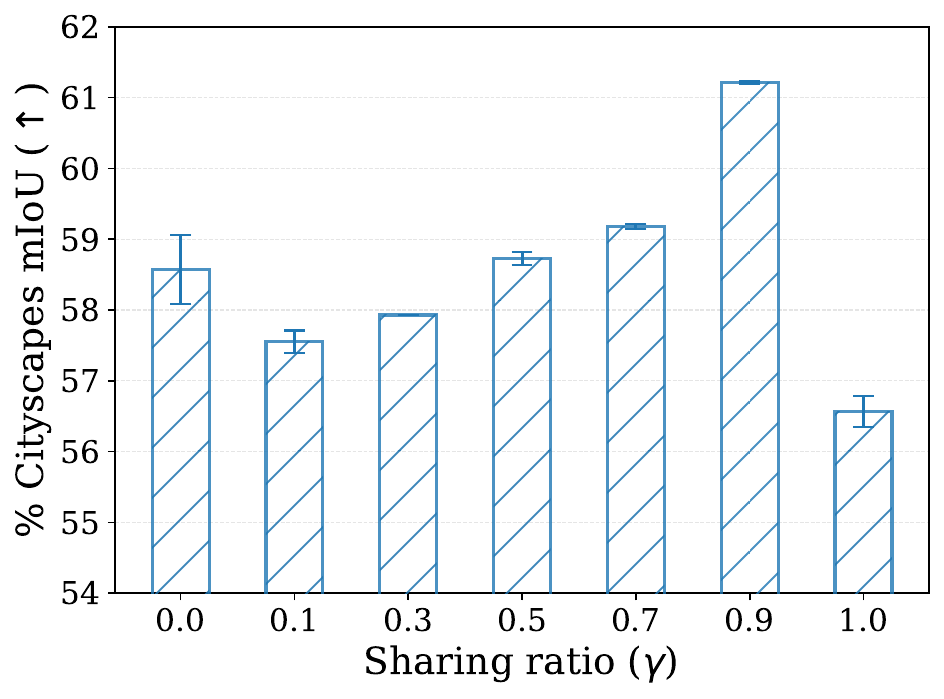}
    \hfill
    \includegraphics[width=0.495\textwidth]{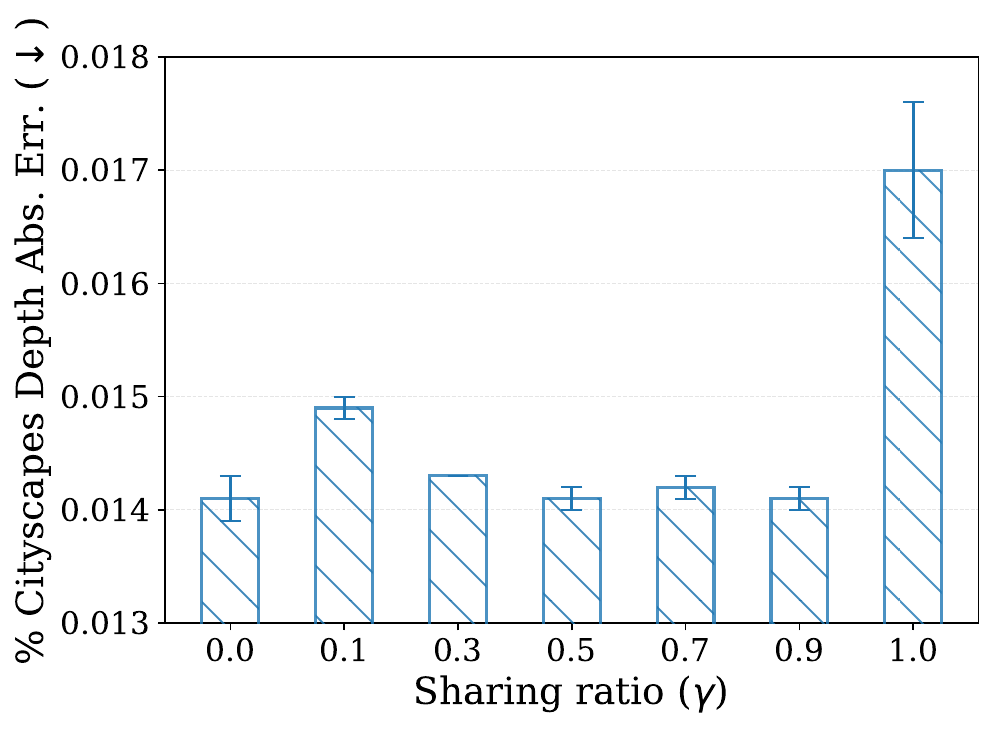}
    \caption{\scriptsize Cityscapes Pixel-Level Prediction\label{fig:city_gamma}}
    \end{subfigure} 
    \vspace{-2em}
    \caption{Effects of sharing ratio $\gamma$ on image-level classification and pixel-level prediction MTL problems.\label{fig:abl_gamma}\vspace{-0.1cm}}
\end{figure}

Figure~\ref{fig:abl_gamma} shows the effect of sharing ratio $\gamma$ on explicit task routing (\tb{}) performance over the CelebA and Cityscapes datasets. The results show that $\gamma=0.9$, i.e., 90\% of the features are shared among tasks, leads to the best performance across both image-level and pixel-level tasks. These results suggest that both cases benefit by sharing a significant number of parameters while still needing a small fraction of task-specific parameters. It is worth noting that when $\gamma=1$, i.e., all tasks share all the parameters of the multi-task network, the performance is impaired due to mutual interference between tasks. Accordingly, we set $\gamma=0.9$ for all experiments shown in the main results section.

\begin{figure}[ht]
    \begin{subfigure}[b]{0.235\textwidth}
    \centering
    \includegraphics[width=0.95\textwidth]{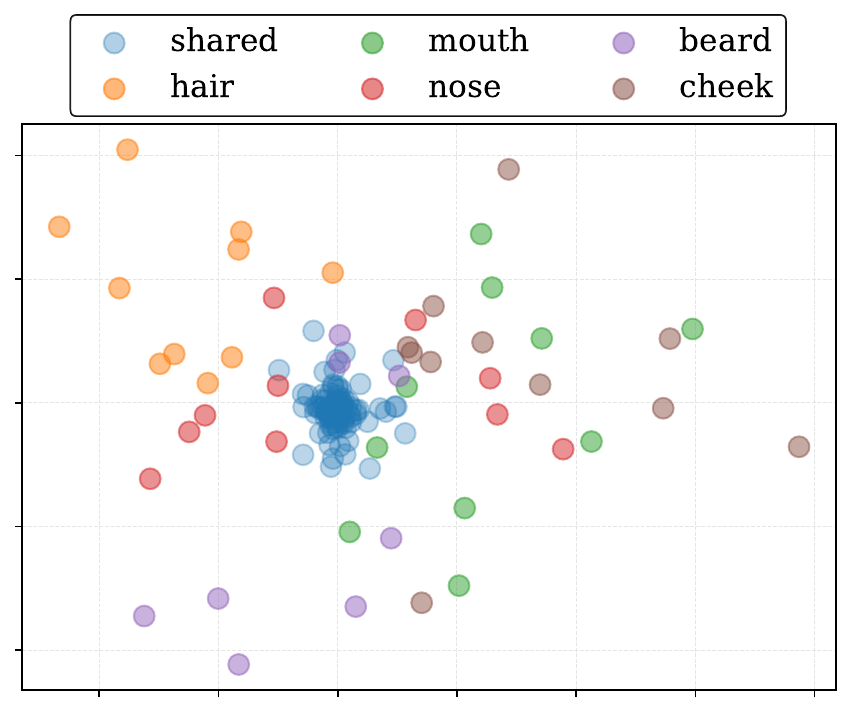}
    \caption{\scriptsize CelebA Image-Level Classification\label{fig:celeba_mb_features}}
    \end{subfigure} \hfill
    \begin{subfigure}[b]{0.235\textwidth}
    \centering
    \includegraphics[width=0.95\textwidth]{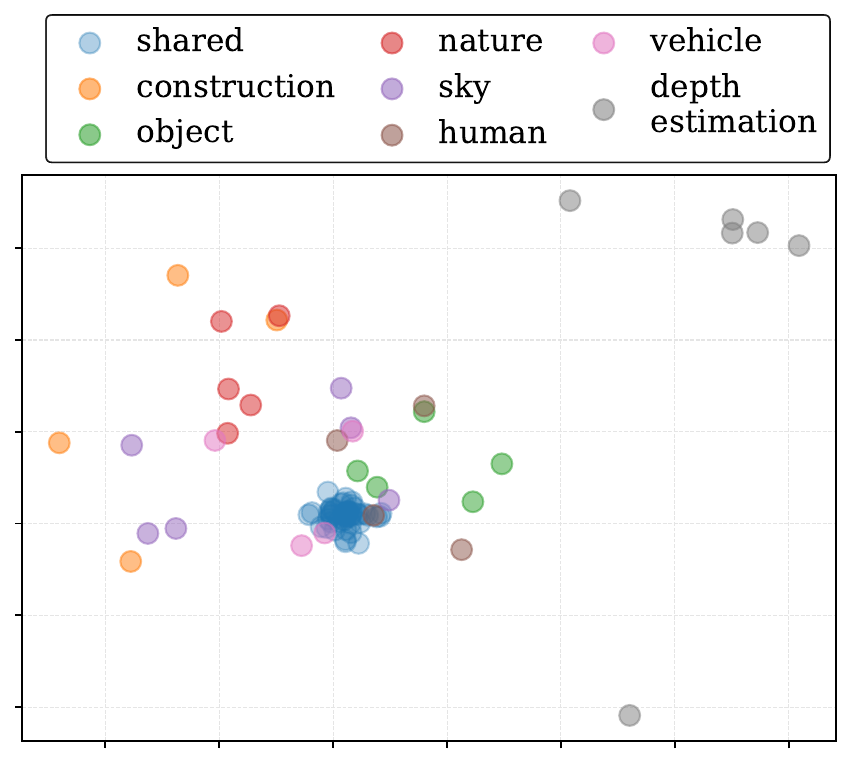}
    \caption{\scriptsize Cityscapes Pixel-Level Prediction\label{fig:city_mb_features}}
    \end{subfigure}
    \vspace{-2em}
    \caption{\textbf{t-SNE visualization of feature activations} from \tb{}'s shared branch and task-specific branches of a single image on (a) CelebA and (b) Cityscapes. Note that features from the shared branch are clustered, while task-specific branches are spread-out.\label{fig:mb_visualization}\vspace{-0.5cm}}
\end{figure}

\begin{table}[ht]
\centering
\caption{Comparison of \tb{} with baselines on image-level classification and pixel-level prediction MTL problems. Our results are highlighted with shading.\label{tab:abl_mb}}
\vspace{-.7em}
\begin{subtable}[h]{0.47\textwidth}
\caption{{\footnotesize Image-Level Classification}}
\resizebox{\textwidth}{!}{%
\begin{tabular}{@{}llcccr@{\hspace{2mm}}}
\toprule
 & Method (ResNet18) & $^{\#}$P (M) & Prec. ($\uparrow$) & Recall ($\uparrow$) & $\Delta_{\rm{p}}$ ($\uparrow$) \\ \midrule
\parbox[t]{2mm}{\multirow{5}{*}{\rotatebox[origin=c]{90}{CelebA}}} & Hard sharing & 11.2 & $67.7_{\pm{0.8}}$ & $59.8_{\pm{0.3}}$ & $0.0\%$ \\
 & Attentive hard sharing \cite{maninis2019attentive} & 12.9 & $71.1_{\pm{0.3}}$ & $62.6_{\pm{0.5}}$ & \textcolor{better}{+$4.9\%$} \\
 & Task routing ($\gamma=0.9$) \cite{Strezoski@Many} & 11.2 & $71.7_{\pm{0.1}}$ & $61.7_{\pm{0.5}}$ & \textcolor{better}{+$4.5\%$} \\
 & Max roaming ($\gamma=0.9$) \cite{pascal2020maximum} & 11.2 & $71.2_{\pm{0.4}}$ & $63.0_{\pm{0.6}}$ & \textcolor{better}{+$5.3\%$} \\
 & \cellcolor{gray!20}\tb{} ($\gamma=0.9$) & \cellcolor{gray!20}11.2 & \cellcolor{gray!20}$\bm{72.0_{\pm{0.4}}}$ & \cellcolor{gray!20}$\bm{63.6_{\pm{0.2}}}$ & \cellcolor{gray!20}\textcolor{better}{+$\bm{6.4\%}$}\\ \bottomrule
\end{tabular}%
}
\end{subtable}\\
\vspace{.5em}
\begin{subtable}[h]{0.48\textwidth}
\caption{\footnotesize Pixel-Level Prediction}
\resizebox{\textwidth}{!}{%
\begin{tabular}{@{}llcccr@{\hspace{2mm}}}
\toprule
 & \multirow{2}{*}{\begin{tabular}[l]{@{}l@{}}Method\\(SegNet)\end{tabular}}  & \multirow{2}{*}{\begin{tabular}[c]{@{}c@{}}$^{\#}$P\\ (M)\end{tabular}} & \multirow{2}{*}{\begin{tabular}[c]{@{}c@{}}Segm.\\ mIoU ($\uparrow$)\end{tabular}} & \multirow{2}{*}{\begin{tabular}[c]{@{}c@{}}Depth\\ Abs. Err. ($\downarrow$)\end{tabular}} & \multirow{2}{*}{\begin{tabular}[c]{@{}c@{}} $\Delta_{\rm{p}}$ ($\uparrow$) \end{tabular}} \\ 
 &  &  &  & \\ \midrule
\parbox[t]{2mm}{\multirow{5}{*}{\rotatebox[origin=c]{90}{Cityscapes}}} & Hard sharing & 25.1 & $56.57_{\pm{0.22}}$ & $0.0170_{\pm{0.0006}}$ & $0.0\%$ \\
 & Attentive hard sharing \cite{maninis2019attentive} & 28.2 & $55.45_{\pm{1.03}}$ & $0.0160_{\pm{0.0006}}$ & \textcolor{better}{+$2.0\%$} \\
 & Task routing ($\gamma=0.6$) \cite{Strezoski@Many} & 25.1 & $56.52_{\pm{0.41}}$ & $0.0155_{\pm{0.0003}}$ & \textcolor{better}{+$4.4\%$} \\
 & Max roaming ($\gamma=0.6$) \cite{pascal2020maximum} & 25.1 & $57.93_{\pm{0.20}}$ & $0.0143_{\pm{0.0001}}$ & \textcolor{better}{+$9.1\%$} \\
 & \cellcolor{gray!20}\tb{} ($\gamma=0.9$) & \cellcolor{gray!20}25.1 & \cellcolor{gray!20}$\bm{61.22_{\pm{0.16}}}$ & \cellcolor{gray!20}$\bm{0.0141_{\pm{0.0001}}}$ & \cellcolor{gray!20}\textcolor{better}{+$\bm{12.6\%}$} \\ \bottomrule
\end{tabular}%
}
\end{subtable}
\end{table}
Table~\ref{tab:abl_mb} compares our explicit task routing with other parameter partitioning methods. All methods use regular convolution. We make the following observations, (i) All parameter partitioning methods improve performance over hard sharing. (ii) Our explicit task routing strategy is more effective for alleviating task interference and improving performance on both image-level classification and pixel-level dense prediction tasks. Additionally, to further understand the utility of explicit task routing, we visualize the features extracted by the shared branch and task-specific branches using t-SNE~\cite{van2008visualizing} in Figure~\ref{fig:mb_visualization}. We observe that the shared branch extracts similar features across tasks, while the task-specific stems extract individualized features for specific tasks. We also notice that the features extracted by task-specific branches are dissimilar for different tasks, which shows that our explicit task routing can obtain task-specific features and mitigate mutual interference between tasks.

\begin{table*}[ht]
\centering
\caption{Comparison of \ourmethod{} with baselines on CelebA image-level classification problems. Our results are highlighted with shading. \label{tab:main_celeba}}
\vspace{-1em}
\resizebox{.80\textwidth}{!}{%
\begin{tabular}{@{\hspace{2mm}}lc|cccr|cccr@{\hspace{2mm}}}
\toprule
\multirow{2}{*}{\begin{tabular}[l]{@{}l@{}}Method\\(ResNet18)\end{tabular}} & \multirow{2}{*}{\begin{tabular}[c]{@{}c@{}}$^{\#}$P\\ (M)\end{tabular}} & \multicolumn{4}{c}{8 grouped facial attributes (tasks)} & \multicolumn{4}{|c}{40 facial attributes (tasks)} \\ \cmidrule(lr){3-6} \cmidrule(lr){7-10}
 &  & Precision ($\uparrow$) & Recall ($\uparrow$) & F-score ($\uparrow$) & $\Delta_{\rm{p}}$ ($\uparrow$) & Precision ($\uparrow$) & Recall ($\uparrow$) & F-score ($\uparrow$) & $\Delta_{\rm{p}}$ ($\uparrow$) \\ \midrule
Hard sharing & 11.2 & $67.7_{\pm{0.8}}$ & $59.8_{\pm{0.3}}$ & $63.0_{\pm0.2}$ & $0.0\%$ & $70.8_{\pm0.9}$ & $60.0_{\pm0.3}$ & $64.2_{\pm0.1}$ & $0.0\%$\\
GradNorm ($\alpha=0.5$) \cite{chen2018gradnorm} & 11.2 & $70.4_{\pm{0.1}}$ & $59.5_{\pm{0.6}}$ & $63.6_{\pm{0.5}}$ & \textcolor{better}{+$1.5\%$} & $70.7_{\pm{0.8}}$ & $60.0_{\pm{0.3}}$ & $64.1_{\pm{0.3}}$ & \textcolor{red}{-$0.1\%$} \\
MGDA-UB \cite{MGDA} & 11.2 & $68.6_{\pm{0.1}}$ & $60.2_{\pm{0.3}}$ & $63.6_{\pm{0.3}}$ & \textcolor{better}{+$1.0\%$} & $71.8_{\pm{0.9}}$ & $57.4_{\pm{0.3}}$ & $62.3_{\pm{0.2}}$\ & \textcolor{red}{-$2.0\%$} \\
Atten. hard sharing \cite{maninis2019attentive} & 12.9 & $71.1_{\pm{0.3}}$ & $62.6_{\pm{0.5}}$ & $65.9_{\pm{0.2}}$ & \textcolor{better}{+$4.8\%$} & $73.2_{\pm{0.1}}$ & $63.6_{\pm{0.2}}$ & $67.5_{\pm{0.1}}$ & \textcolor{better}{+$4.8\%$}\\
Task routing \cite{Strezoski@Many} & 11.2 & $71.7_{\pm{0.1}}$ & $61.7_{\pm{0.5}}$ & $65.5_{\pm{0.3}}$ & \textcolor{better}{+$4.4\%$} & $72.1_{\pm{0.8}}$ & $63.4_{\pm{0.3}}$ & $66.8_{\pm{0.2}}$ & \textcolor{better}{+$3.9\%$}\\
Max roaming \cite{pascal2020maximum} & 11.2 & $71.2_{\pm{0.4}}$ & $63.0_{\pm{0.6}}$ & $66.2_{\pm{0.2}}$ & \textcolor{better}{+$5.2\%$} & $73.0_{\pm{0.4}}$ & $63.6_{\pm{0.1}}$ & $67.3_{\pm{0.1}}$ & \textcolor{better}{+$4.6\%$}\\
\rowcolor{gray!20}
\ourmethod{} & 8.0 & $\bm{72.7_{\pm{0.4}}}$ & $\bm{64.8_{\pm{0.3}}}$ & $\bm{67.8_{\pm{0.1}}}$ & \textcolor{better}{+$\bm{7.8\%}$} & $\bm{73.2_{\pm{0.2}}}$ & $\bm{64.8_{\pm{0.3}}}$ & $\bm{68.1_{\pm{0.1}}}$ & \textcolor{better}{+$\bm{5.8\%}$} \\
\bottomrule
\end{tabular}%
}
\end{table*}
\subsubsection{\ourmethod{} Networks\label{sec:main_results}}
Table~\ref{tab:main_celeba} presents the performance of various methods on the CelebA image-level classification problems. We consider two experimental settings, one with eight group-level tasks and another with 40 binary tasks. We make the following observations, (i) Our \ourmethod{} is consistently better than the baselines methods while having fewer learnable parameters. (ii) We observe that the performance gains become more prominent as the number of tasks increases due to the inherent minimization of interference between tasks in \ourmethod{}. (iii) We also observe that parameter partitioning methods (i.e., task routing, max roaming, and \ourmethod{}) scale better than loss/gradient balancing methods (i.e., GradNorm and MGDA-UB) to a higher number of tasks. For instance, when the number of tasks increases from 8 to 40, the F-score of MGDA-UB decreases by 1.3\%, while the F-score of our proposed \ourmethod{} improves by 0.3\% while having 28.5\% fewer (11.2M to 8M) learnable parameters.

Table~\ref{tab:main_nyuv2} and Table~\ref{tab:main_cityscapes} show the experimental results for pixel-level dense prediction problems on NYUv2 and Cityscapes datasets, respectively. Again, we observe that \ourmethod{} significantly outperforms the baselines. Furthermore, it is worth mentioning that our \ourmethod{} is the highest on all metrics. For instance, as shown in Table~\ref{tab:main_cityscapes}, \ourmethod{} obtains 61.49 mIoU for semantic segmentation, an improvement of +3.56 mIoU over the previous state-of-the-art results, while having 13.5\% less learnable parameters. A similarly significant improvement is observed on the NYU-v2 dataset across \emph{all} tasks.

\begin{table*}[!ht]
\centering
\caption{Comparison of \ourmethod{} with baselines on NYU-v2 pixel-level prediction problems. Our results are highlighted with shading. \label{tab:main_nyuv2}}
\resizebox{.95\textwidth}{!}{%
\begin{tabular}{@{\hspace{2mm}}lcccccccccr@{\hspace{2mm}}}
\toprule
\multirow{3}{*}{\begin{tabular}[l]{@{}l@{}}Method\\(SegNet)\end{tabular}} & \multirow{3}{*}{\begin{tabular}[c]{@{}c@{}}$^{\#}$P\\ (M)\end{tabular}} & \multirow{3}{*}{\begin{tabular}[c]{@{}c@{}}Segm. \\ mIoU ($\uparrow$)\end{tabular}} & \multicolumn{2}{c}{Depth Estimation} & \multicolumn{5}{c}{Surface Normal Estimation} & 
\multirow{3}{*}{\begin{tabular}[c]{@{}c@{}}$\Delta_{\rm{p}}$ ($\uparrow$)\end{tabular}} \\
 &  &  & \multicolumn{2}{c}{(Lower better $\downarrow$)} & \multicolumn{2}{c}{Angle distance ($\downarrow$)} & \multicolumn{3}{c}{Within $t^{o}$ ($\uparrow$)} & \\\cmidrule(lr){4-5} \cmidrule(lr){6-7} \cmidrule(lr){8-10}
 &  &  & Abs. Err. & Rel. Err.  & Mean Err. & Median Err. & $11.25$ & $22.5$ & $30$ & \\ \midrule
Hard sharing & 25.1 & $15.98_{\pm{0.56}}$ & $0.6095_{\pm{0.0041}}$ & $0.2554_{\pm 0.0007}$ & $32.43_{\pm{0.19}}$ & $27.43_{\pm{0.35}}$ & $20.66_{\pm{0.19}}$ & $42.84_{\pm{0.19}}$ & $55.02_{\pm{0.19}}$ & $0.0\%$ \\
GradNorm \cite{chen2018gradnorm} & 25.1 & $16.13_{\pm{0.23}}$ & $0.7626_{\pm{0.0034}}$ & $0.3208_{\pm 0.0050}$ & $34.45_{\pm{0.52}}$ & $30.98_{\pm{0.80}}$ & $18.96_{\pm{0.60}}$ & $40.85_{\pm{0.92}}$ & $53.34_{\pm{0.24}}$ & \textcolor{red}{-$10.6\%$}\\
Cross-Stitch \cite{Misra@Cross} & 75.3 & $14.71_{\pm{0.23}}$ & $0.6481_{\pm{0.0034}}$ & $0.2871_{\pm 0.0050}$ & $33.56_{\pm{0.52}}$ & $28.58_{\pm{0.80}}$ & $20.08_{\pm{0.80}}$ & $40.54_{\pm{0.80}}$ & $51.97_{\pm{0.80}}$ & \textcolor{red}{-$6.0\%$} \\
MTAN \cite{liu2019end} & 44.4 & $17.72_{\pm{0.23}}$ & $0.5960_{\pm{0.0034}}$ & $0.2577_{\pm 0.0050}$ & $31.44_{\pm{0.52}}$ & $25.37_{\pm{0.80}}$ & $23.17_{\pm{0.80}}$ & $45.65_{\pm{0.80}}$ & $57.48_{\pm{0.80}}$ & \textcolor{better}{+$5.7\%$}\\
Atten. \cite{maninis2019attentive} & 25.1 & $16.02_{\pm{0.12}}$ & $0.5988_{\pm{0.0112}}$ & $0.2630_{\pm 0.0058}$ & $32.22_{\pm{0.02}}$ & $26.12_{\pm{0.02}}$ & $20.44_{\pm{0.09}}$ & $42.86_{\pm{0.34}}$ & $55.14_{\pm{0.67}}$ & \textcolor{better}{+$0.5\%$}\\
Task routing \cite{Strezoski@Many} & 25.1 & $16.54_{\pm{0.02}}$ & $0.6354_{\pm{0.0085}}$ & $0.2786_{\pm 0.0090}$ & $30.93_{\pm{0.19}}$ & $25.51_{\pm{0.28}}$ & $22.52_{\pm{0.36}}$ & $45.41_{\pm{0.82}}$ & $57.46_{\pm{0.37}}$ & \textcolor{better}{+$2.7\%$}\\
Max roaming \cite{pascal2020maximum} & 25.1 & $17.40_{\pm{0.31}}$ & $0.6082_{\pm{0.0023}}$ & $0.2750_{\pm 0.0015}$ & $30.58_{\pm{0.04}}$ & $24.67_{\pm{0.08}}$ & $23.74_{\pm{0.61}}$ & $46.75{\pm{0.41}}$ & $58.84_{\pm{0.28}}$ & \textcolor{better}{+$6.0\%$}\\
\rowcolor{gray!20}
\ourmethod{} & 25.1 & $\bm{20.37_{\pm{0.32}}}$ & $\bm{0.5790_{\pm{0.0067}}}$ & $\bm{0.2510_{\pm 0.0090}}$ & $\bm{28.92_{\pm{0.05}}}$ & $\bm{23.22_{\pm{0.16}}}$ & $\bm{25.38_{\pm{0.11}}}$ & $\bm{49.11_{\pm{0.27}}}$ & $\bm{61.22_{\pm{0.23}}}$ & \textcolor{better}{+$\bm{13.6\%}$}\\
\bottomrule
\end{tabular}%
}
\end{table*}

\begin{table}[!ht]
\centering
\caption{Comparison of \ourmethod{} with baselines on Cityscapes pixel-level predictions. Our results are highlighted with shading.\label{tab:main_cityscapes}}
\vspace{-.5em}
\resizebox{.485\textwidth}{!}{%
\begin{tabular}{@{\hspace{2mm}}lccccr@{\hspace{2mm}}}
\toprule
\multirow{2}{*}{\begin{tabular}[l]{@{}l@{}}Method\\(SegNet)\end{tabular}} & \multirow{2}{*}{\begin{tabular}[c]{@{}c@{}}$^{\#}$P\\ (M)\end{tabular}} & \multirow{2}{*}{\begin{tabular}[c]{@{}c@{}}Segm. \\ mIoU ($\uparrow$)\end{tabular}} & \multicolumn{2}{c}{Depth Estimation} & \multirow{2}{*}{\begin{tabular}[c]{@{}c@{}}$\Delta_{\rm{p}}$ ($\uparrow$)\end{tabular}} \\ \cmidrule(lr){4-5} 
 &  &  & Abs. Err. ($\downarrow$) & Rel. Err. ($\downarrow$) &  \\ \midrule
Hard sharing & 25.1 & $56.57_{\pm{0.22}}$ & $0.0170_{\pm{0.0006}}$ & $43.99_{\pm5.53}$ & $0.0\%$  \\
GradNorm \cite{chen2018gradnorm} & 25.1 & $56.77_{\pm{0.08}}$ & $0.0199_{\pm{0.0004}}$ & $68.13_{\pm4.48}$ & \textcolor{red}{-$23.9\%$} \\
Cross-Stitch \cite{Misra@Cross} & 75.3 & $50.08_{\pm{0.23}}$ & $0.0154_{\pm{0.0001}}$ & $34.49_{\pm 1.24}$ & \textcolor{better}{+$6.5\%$} \\
MTAN \cite{liu2019end} & 44.4 & $53.04_{\pm{0.32}}$ & $0.0144_{\pm{0.0001}}$ & $33.63_{\pm 1.51}$ & \textcolor{better}{+$10.9\%$} \\
Atten. \cite{maninis2019attentive} & 25.1 & $55.45_{\pm{1.03}}$ & $0.0160_{\pm{0.0006}}$ & $35.72_{\pm 1.62}$ & \textcolor{better}{+$7.6\%$} \\
Task routing \cite{Strezoski@Many} & 25.1 & $56.52_{\pm{0.41}}$ & $0.0155_{\pm{0.0003}}$ & $31.47_{\pm 0.55}$ & \textcolor{better}{+$12.4\%$}\\
Max roaming \cite{pascal2020maximum} & 25.1 & $57.93_{\pm{0.20}}$ & $0.0143_{\pm{0.0001}}$ & $29.38_{\pm 1.66}$ & \textcolor{better}{+$17.2\%$}\\
\rowcolor{gray!20}
\ourmethod{} & 22.1 & $\bm{61.49_{\pm{0.29}}}$ & $\bm{0.0136_{\pm{0.0001}}}$ & $\bm{29.16_{\pm 1.30}}$ & \textcolor{better}{+$\bm{20.8\%}$}\\
\bottomrule
\end{tabular}%
}
\end{table}
\begin{table}[ht]
\centering
\caption{Comparison of different training strategies for \tb{} on image-level classification and pixel-level prediction problems.}
\label{tab:my-table}
\vspace{-.5em}
\resizebox{.45\textwidth}{!}{%
\begin{tabular}{@{\hspace{2mm}}lcccc@{\hspace{2mm}}}
\toprule
\multirow{2}{*}{Method} & \multicolumn{2}{c}{CelebA} & \multicolumn{2}{c}{Cityscapes} \\ \cmidrule(lr){2-3} \cmidrule(lr){4-5} 
 & Precision ($\uparrow$) & Recall ($\uparrow$) & mIoU ($\uparrow$) & Abs. Err. ($\downarrow$) \\ \midrule
steady-state & ${72.0_{\pm{0.4}}}$ & ${63.6_{\pm{0.2}}}$ & ${59.70_{\pm{0.38}}}$ & ${0.0139_{\pm{0.0002}}}$ \\
synchronized & ${23.5_{\pm{2.4}}}$ & ${42.1_{\pm{3.7}}}$ & ${61.22_{\pm{0.16}}}$ & ${0.0141_{\pm{0.0001}}}$ \\ \bottomrule
\end{tabular}%
}
\end{table}
\begin{figure*}[!ht]
    \centering
    \begin{subfigure}[b]{0.16\linewidth}
        \centering
        \includegraphics[height=.7\linewidth]{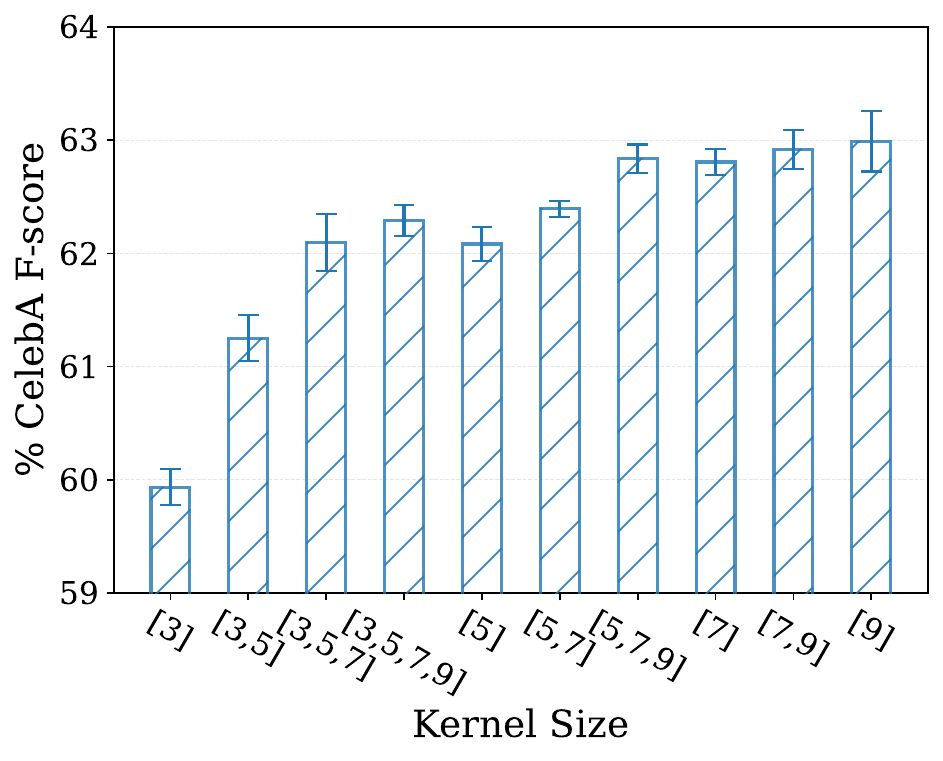}
        \caption{\scriptsize Avg. Pooling \label{fig:abl_nl_avg_pool_celeba}}
    \end{subfigure}
    \begin{subfigure}[b]{0.16\linewidth}
        \centering
        \includegraphics[height=.7\linewidth]{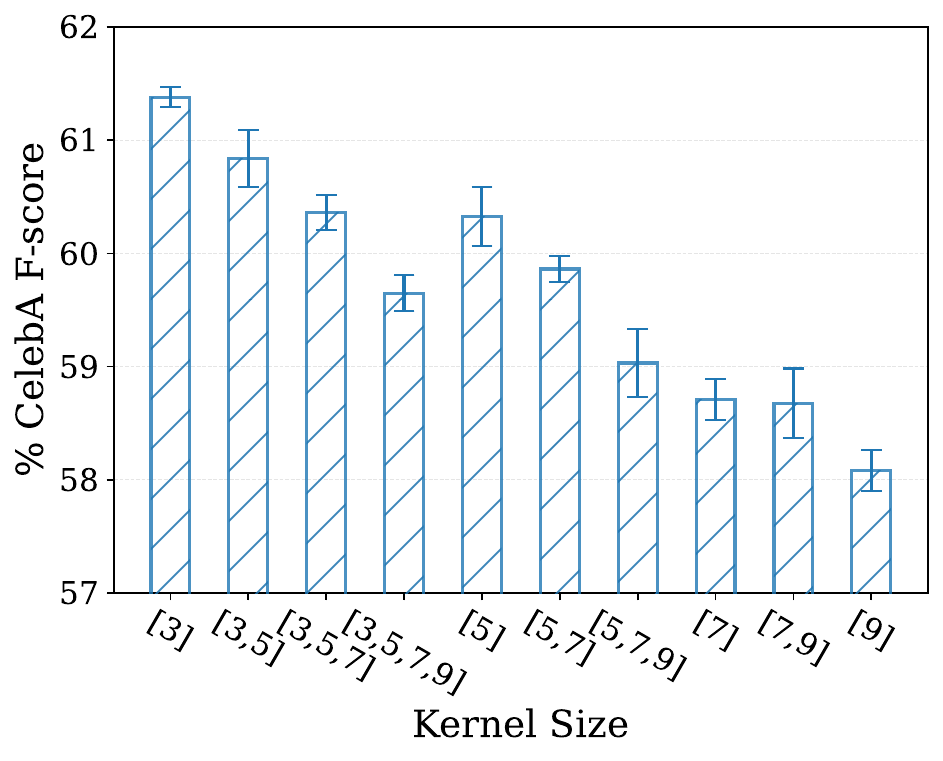}
        \caption{\scriptsize Max Pooling \label{fig:abl_nl_max_pool_celeba}}
    \end{subfigure}
    \begin{subfigure}[b]{0.16\linewidth}
        \centering
        \includegraphics[height=.7\linewidth]{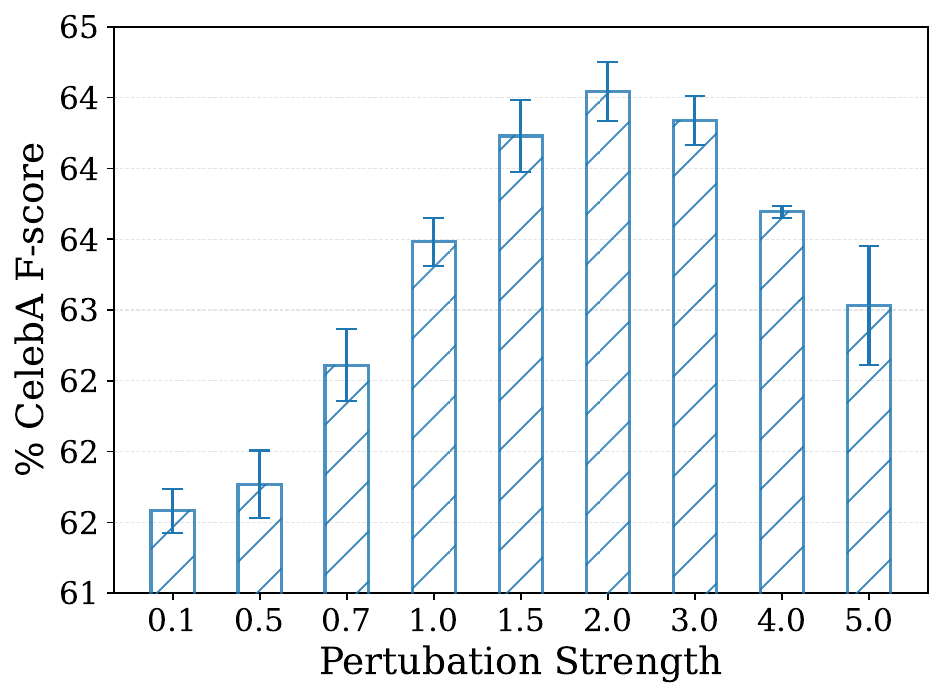}
        \caption{\scriptsize Perturbation \label{fig:abl_nl_perturbation_celeba}}
    \end{subfigure}
    \begin{subfigure}[b]{0.16\linewidth}
        \centering
        \includegraphics[height=.7\linewidth]{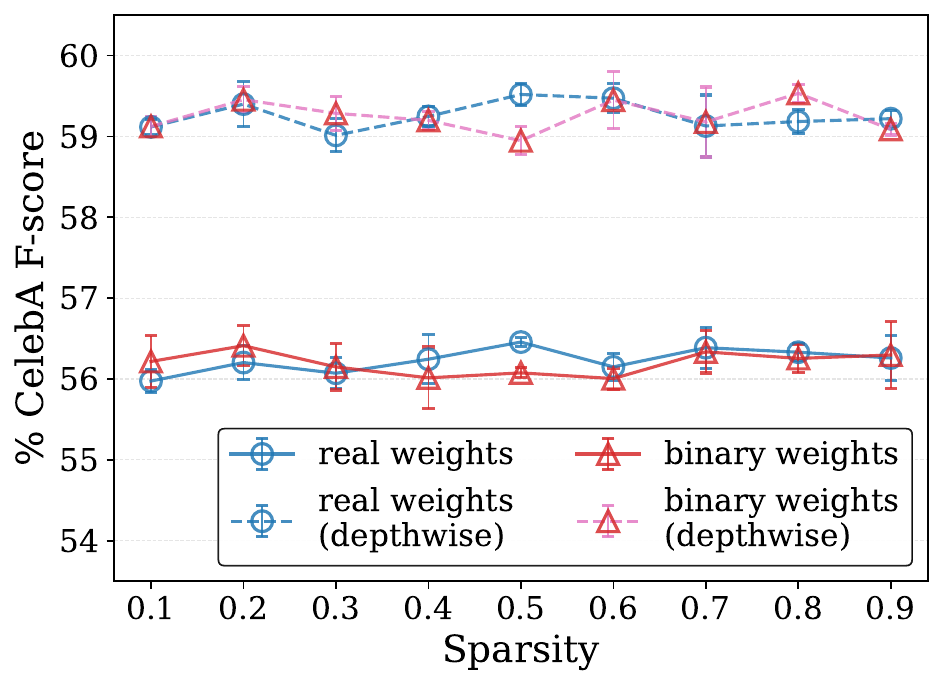}
        \caption{\scriptsize Convolution (sparsity) \label{fig:abl_nl_conv_depth_celeba}}
    \end{subfigure}
    \begin{subfigure}[b]{0.34\linewidth}
        \centering
        \includegraphics[height=.34\linewidth]{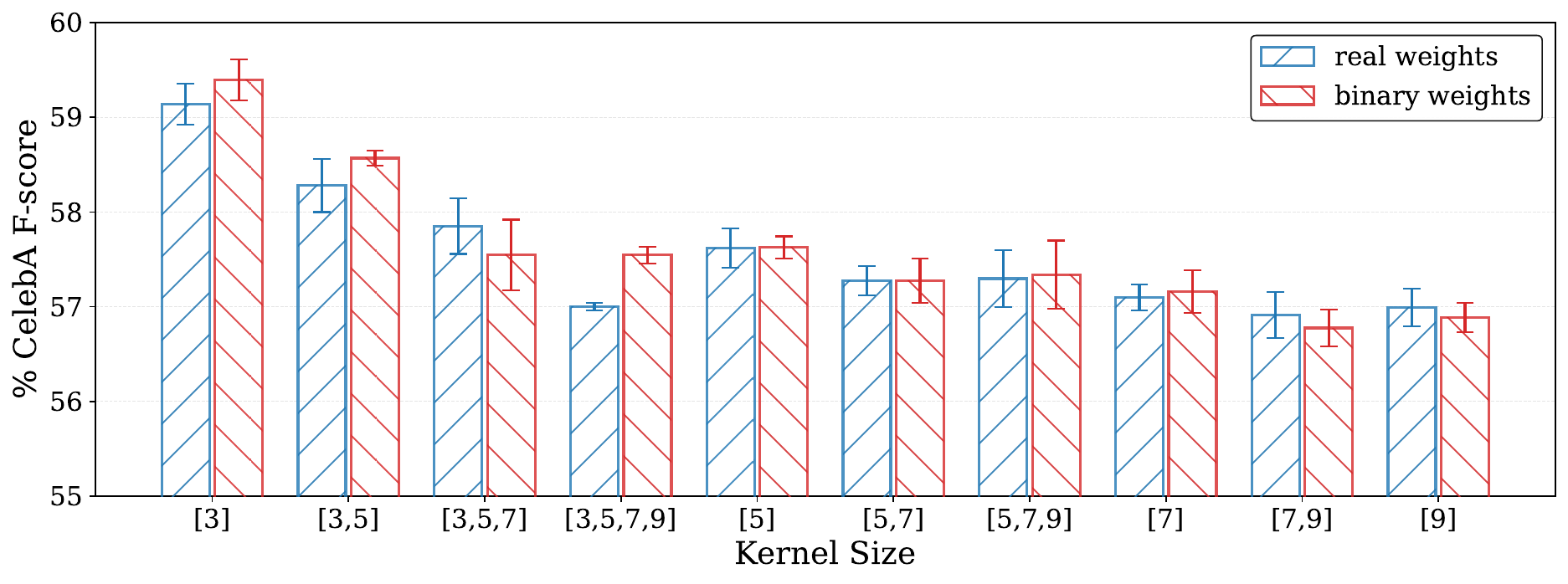}
        \caption{\scriptsize Convolution (kernel size) \label{fig:abl_nl_conv_ks_celeba}}
    \end{subfigure}
    \vspace{-2em}
    \caption{Effect of different hyperparameters of individual NLP.\label{fig:abl_nl_layer_hyper}\vspace{-0.2cm}}
\end{figure*}
\section{Ablation Analysis\label{sec:ablation}}
\noindent\textbf{NLP Hyperparameters:} Figure~\ref{fig:abl_nl_layer_hyper} shows the effect of different settings for individual NLPs for image-level classification tasks. We observe that (i) A combination of different kernel sizes can improve performance even for the same type of NLP (e.g., avg pooling), as kernels of different sizes can extract diverse features. These results also indicate that multi-task learning benefits from operating on diverse features. (ii) For each type ofLP, the choice of parameters greatly impacts performance. For instance, for avg pooling, the F-score with a kernel size of 9 is 3\% higher than that with a kernel size of 3.

The final design of our proposed NLP for image-level classification tasks was guided by the observations that we summarize as follows: (1) average pooling with larger kernels outperforms max pooling; (2) depth-wise convolutions outperform full convolutions while there is no appreciable difference between using real-valued or binary weights; (3) smaller convolution kernels outperform larger ones; (4) perturbation helps improve performance.

\vspace{3pt}
\noindent\textbf{Training strategy for \ourmethod{}:} Table~\ref{tab:my-table} shows the experimental results of ``Steady-state'' and ``Synchronized'' training strategies on the CelebA image-level and Cityscapes pixel-level prediction tasks. ``Steady-state'' refers to updating the parameters of the shared branch immediately after forwarding on one task, while ``Synchronized'' refers to waiting until all tasks are forwarded before updating the parameters of the shared branch. For image-level classification tasks, ``steady-state'' training is better, while for pixel-level prediction tasks, ``synchronized'' training is better.

%% file: 5-conclusion.tex
\section{Conclusion}
In this paper, we present the Explicit Task Routing with Non-Learnable Primitives (\ourmethod{}) module for multi-task learning. The \ourmethod{} module introduces non-learnable primitives for extracting task-agnostic and diverse features and explicit task routing to extract task-specific features for each task. Both non-learnable primitives and explicit task routing can provide the flexibility needed to minimize task interference. Experiments on the CelebA dataset with multiple image-level classification tasks and on the NYU-v2 and Cityscapes datasets with multiple pixel-level prediction tasks show that our \ourmethod{} method significantly outperforms state-of-the-art baselines with fewer learnable parameters and similar FLOPs across all datasets.

%% file: 6-appendix.tex
\noindent This appendix includes the following:
\begin{enumerate}
    \item Extended description of Related Work in Section~\ref{sec:app_related_work}.
    \item Additional results of configurations on NLPs-based feature extraction in Section~\ref{sec:app_nlp_results}.
    \item Extended description of datasets and baseline methods in Section~\ref{sec:app_datasets_baselines}. 
\end{enumerate}

\section{An Extended Description of Related Work \label{sec:app_related_work}}

Existing methods related to MTL architectures can be classified into encoder or decoder-focused ones. Encoder-focused approaches primarily lay emphasis on architectures that can encode multi-purpose feature representations through supervision from multiple tasks. Such encoding is typically achieved, for example, via feature fusion~\cite{Misra@Cross,Gao@NDDR,Ma@SNR}, branching~\cite{Kokkinos@UberNet,Lu@Fully,Vandenhende@Branched,Long@Learning}, self-supervision~\cite{doersch2017multi}, shared and task-specific modules~\cite{liu2019end,maninis2019attentive}, filter grouping~\cite{Bragman@Stochastic}, filter modulation~\cite{Kanakis@Reparameterizing,Zhao@AModulation}, task routing~\cite{Strezoski@Many,Ma@SNR,Rosenbaum@Routing}, or neural architecture search~\cite{Gao@MTLNAS}. Decoder-focused approaches start from the feature representations learned at the encoding stage, and further refine them at the decoding stage by distilling information across tasks in a one-off~\cite{Xu@PAD}, sequential~\cite{zhang2018joint}, recursive~\cite{Zhang@Pattern}, or even multi-scale~\cite{vandenhende2020mti} manner. Due to inherent layer sharing, the approaches above typically suffer from task interference and negative transfer~\cite{Wu@Understanding}.

In the context of MTL, our explicit task routing layer is conceptually related to~\cite{liu2019end,maninis2019attentive}, but notably different in motivation and design. First, both of these two approaches operate on the features obtained from a shared backbone to extract task-specific features. In contrast, our shared and task-specific branches operate in parallel and extract features from the common features extracted by the non-learnable layer. Second, these existing approaches utilize attention mechanisms to distill task-specific features from the shared features, while we use lightweight $1\times 1$ convolution for the same purpose. Third, our explicit task routing layer is tailored to exploit the non-learnable layer for MTL optimally. Finally, unlike baselines, our multi-branch design affords simple and explicit control over the ratio of shared and task-specific parameters.

Additionally, our work is also closely related to reparameterized convolutions for multi-task learning (RCM)~\cite{Kanakis@Reparameterizing}, which first introduced the concept of using non-learnable convolutional filters for MTL. However, there are three notable differences. First, the non-learnable layer of RCM only includes standard convolution, while we consider other non-learnable primitives such as pooling, identity, and additive noise \cite{XU@Perturbative} operations. Second, RCM uses pre-trained network weights to initialize non-learnable convolutional filters, while in our case they are sampled from a random distribution. Relying on pre-trained weights limits RCM's ability to reduce the model size and its generalizability to architectures without readily available pre-trained weights. Finally, there is no collaboration between tasks in RCM as it only comprises task-specific modulators, while we utilize a shared branch to help tasks use each other's training signals. Having both shared and task-specific branches allows tasks to amortize parameters that are commonly useful across multiple tasks, thereby minimizing redundancy in the task-specific branches, unlike RCM. Moreover, our method also offers fine-grained control over the ratio of parameters that are shared or task-specific. 

\section{Results of NLPs-based feature extraction\label{sec:app_nlp_results}}

In this section, we first provide the full version of Table 1 from the main paper in Table S1, showing the effect of different configurations of \nlp{}s on CelebA multi-attribute classification. 
Then, we present the effect of different hyperparameter settings of \nlp{}s on NYU-v2 dense prediction MTL problem in Figure S1. 

\begin{table}[t]
\centering
\caption*{Table S1. Effect of different configurations of \nlp{}s on CelebA multi-attribute classification.}
\resizebox{.485\textwidth}{!}{%
\begin{tabular}{@{\hspace{2mm}}ccccc|ccc@{\hspace{2mm}}}
\toprule
\multicolumn{5}{c|}{Non-learnable Operators} & \multicolumn{3}{c}{CelebA} \\ \cmidrule(r){1-5} \cmidrule(l){6-8}
Avg. pool & Max pool & Conv & Shift & Noise & Precision & Recall & F-Score \\ \midrule
 &  & \checkmark &  &  & $73.37_{\textcolor{gray}{\pm0.19}}$ & $59.00_{\textcolor{gray}{\pm0.11}}$ & $61.08_{\textcolor{gray}{\pm0.17}}$ \\ 
 & \checkmark & \checkmark &  &  & $72.50_{\textcolor{gray}{\pm0.13}}$ & $57.87_{\textcolor{gray}{\pm0.09}}$ & $61.10_{\textcolor{gray}{\pm0.12}}$ \\ 
 & \checkmark &  &  &  & $72.29_{\textcolor{gray}{\pm0.14}}$ & $58.65_{\textcolor{gray}{\pm0.27}}$ & $61.30_{\textcolor{gray}{\pm0.04}}$ \\ 
\checkmark & \checkmark & \checkmark &  &  & $73.24_{\textcolor{gray}{\pm0.17}}$ & $58.87_{\textcolor{gray}{\pm0.30}}$ & $61.46_{\textcolor{gray}{\pm0.13}}$ \\ 
\checkmark & \checkmark & \checkmark & \checkmark &  & $73.37_{\textcolor{gray}{\pm0.31}}$ & $58.67_{\textcolor{gray}{\pm0.07}}$ & $61.56_{\textcolor{gray}{\pm0.24}}$ \\ 
\checkmark & \checkmark &  &  &  & $72.92_{\textcolor{gray}{\pm0.13}}$ & $58.76_{\textcolor{gray}{\pm0.08}}$ & $61.57_{\textcolor{gray}{\pm0.06}}$ \\ 
\checkmark & \checkmark &  & \checkmark &  & $73.51_{\textcolor{gray}{\pm0.39}}$ & $59.47_{\textcolor{gray}{\pm0.25}}$ & $61.81_{\textcolor{gray}{\pm0.15}}$ \\ 
 & \checkmark &  & \checkmark &  & $72.80_{\textcolor{gray}{\pm0.33}}$ & $59.39_{\textcolor{gray}{\pm0.13}}$ & $61.83_{\textcolor{gray}{\pm0.20}}$ \\ 
 & \checkmark & \checkmark & \checkmark &  & $72.93_{\textcolor{gray}{\pm0.15}}$ & $59.47_{\textcolor{gray}{\pm0.15}}$ & $62.04_{\textcolor{gray}{\pm0.20}}$ \\ 
 &  &  & \checkmark &  & $73.56_{\textcolor{gray}{\pm0.18}}$ & $59.90_{\textcolor{gray}{\pm0.14}}$ & $62.16_{\textcolor{gray}{\pm0.07}}$ \\ 
 &  & \checkmark & \checkmark &  & $73.52_{\textcolor{gray}{\pm0.62}}$ & $59.16_{\textcolor{gray}{\pm0.08}}$ & $62.19_{\textcolor{gray}{\pm0.24}}$ \\ 
\checkmark &  &  & \checkmark &  & $74.23_{\textcolor{gray}{\pm0.48}}$ & $59.68_{\textcolor{gray}{\pm0.17}}$ & $62.36_{\textcolor{gray}{\pm0.13}}$ \\ 
\checkmark &  & \checkmark & \checkmark &  & $73.97_{\textcolor{gray}{\pm0.47}}$ & $59.85_{\textcolor{gray}{\pm0.17}}$ & $62.41_{\textcolor{gray}{\pm0.08}}$ \\ 
\checkmark &  & \checkmark &  &  & $74.81_{\textcolor{gray}{\pm0.39}}$ & $59.84_{\textcolor{gray}{\pm0.29}}$ & $62.84_{\textcolor{gray}{\pm0.11}}$ \\ 
\checkmark &  &  &  &  & $74.75_{\textcolor{gray}{\pm0.36}}$ & $60.49_{\textcolor{gray}{\pm0.05}}$ & $63.03_{\textcolor{gray}{\pm0.32}}$ \\
\checkmark & \checkmark & \checkmark & \checkmark & \checkmark & $74.56_{\textcolor{gray}{\pm0.24}}$ & $61.24_{\textcolor{gray}{\pm0.61}}$ & $64.08_{\textcolor{gray}{\pm0.21}}$ \\
\checkmark & \checkmark & \checkmark &  & \checkmark & $74.38_{\textcolor{gray}{\pm0.10}}$ & $61.13_{\textcolor{gray}{\pm0.10}}$ & $64.14_{\textcolor{gray}{\pm0.02}}$ \\ 
\checkmark & \checkmark &  & \checkmark & \checkmark & $74.24_{\textcolor{gray}{\pm0.08}}$ & $61.14_{\textcolor{gray}{\pm0.29}}$ & $64.21_{\textcolor{gray}{\pm0.09}}$ \\ 
 & \checkmark & \checkmark & \checkmark & \checkmark & $74.50_{\textcolor{gray}{\pm0.30}}$ & $61.17_{\textcolor{gray}{\pm0.46}}$ & $64.32_{\textcolor{gray}{\pm0.04}}$ \\ 
 & \checkmark &  & \checkmark & \checkmark & $74.40_{\textcolor{gray}{\pm0.32}}$ & $62.07_{\textcolor{gray}{\pm0.19}}$ & $64.51_{\textcolor{gray}{\pm0.22}}$ \\
 &  &  &  & \checkmark & $75.67_{\textcolor{gray}{\pm0.25}}$ & $59.74_{\textcolor{gray}{\pm0.33}}$ & $64.54_{\textcolor{gray}{\pm0.21}}$ \\ 
\checkmark & \checkmark &  &  & \checkmark & $74.35_{\textcolor{gray}{\pm0.03}}$ & $62.18_{\textcolor{gray}{\pm0.16}}$ & $64.61_{\textcolor{gray}{\pm0.15}}$ \\ 
 & \checkmark & \checkmark &  & \checkmark & $74.91_{\textcolor{gray}{\pm0.14}}$ & $62.16_{\textcolor{gray}{\pm0.34}}$ & $64.65_{\textcolor{gray}{\pm0.05}}$ \\ 
\checkmark &  & \checkmark & \checkmark & \checkmark & $75.35_{\textcolor{gray}{\pm0.10}}$ & $61.99_{\textcolor{gray}{\pm0.20}}$ & $65.03_{\textcolor{gray}{\pm0.02}}$ \\ 
 & \checkmark &  &  & \checkmark & $74.61_{\textcolor{gray}{\pm0.14}}$ & $61.93_{\textcolor{gray}{\pm0.23}}$ & $65.07_{\textcolor{gray}{\pm0.05}}$ \\ 
\checkmark &  &  & \checkmark & \checkmark & $75.31_{\textcolor{gray}{\pm0.22}}$ & $62.89_{\textcolor{gray}{\pm0.21}}$ & $65.42_{\textcolor{gray}{\pm0.15}}$ \\ 
 &  & \checkmark & \checkmark & \checkmark & $75.50_{\textcolor{gray}{\pm0.15}}$ & $62.70_{\textcolor{gray}{\pm0.41}}$ & $65.65_{\textcolor{gray}{\pm0.16}}$ \\ 
 &  &  & \checkmark & \checkmark & $75.74_{\textcolor{gray}{\pm0.29}}$ & $62.80_{\textcolor{gray}{\pm0.19}}$ & $65.81_{\textcolor{gray}{\pm0.13}}$ \\ 
\checkmark &  &  &  & \checkmark & $75.72_{\textcolor{gray}{\pm0.08}}$ & $63.19_{\textcolor{gray}{\pm0.39}}$ & $66.08_{\textcolor{gray}{\pm0.10}}$ \\ 
\checkmark &  & \checkmark &  & \checkmark & $75.82_{\textcolor{gray}{\pm0.29}}$ & $63.19_{\textcolor{gray}{\pm0.36}}$ & $66.26_{\textcolor{gray}{\pm0.25}}$ \\
 &  & \checkmark &  & \checkmark & $76.29_{\textcolor{gray}{\pm0.25}}$ & $62.47_{\textcolor{gray}{\pm0.60}}$ & $66.40_{\textcolor{gray}{\pm0.32}}$ \\ \midrule
\multicolumn{5}{c|}{Standard learnable convolution} & $67.67_{\textcolor{gray}{\pm0.75}}$ & $59.84_{\textcolor{gray}{\pm0.33}}$ & $62.86_{\textcolor{gray}{\pm0.07}}$ \\
\bottomrule
\end{tabular}%
}
\end{table}

\begin{figure}[ht]
    \centering
    \begin{subfigure}[b]{0.485\textwidth}
        \centering
        \includegraphics[width=.325\textwidth]{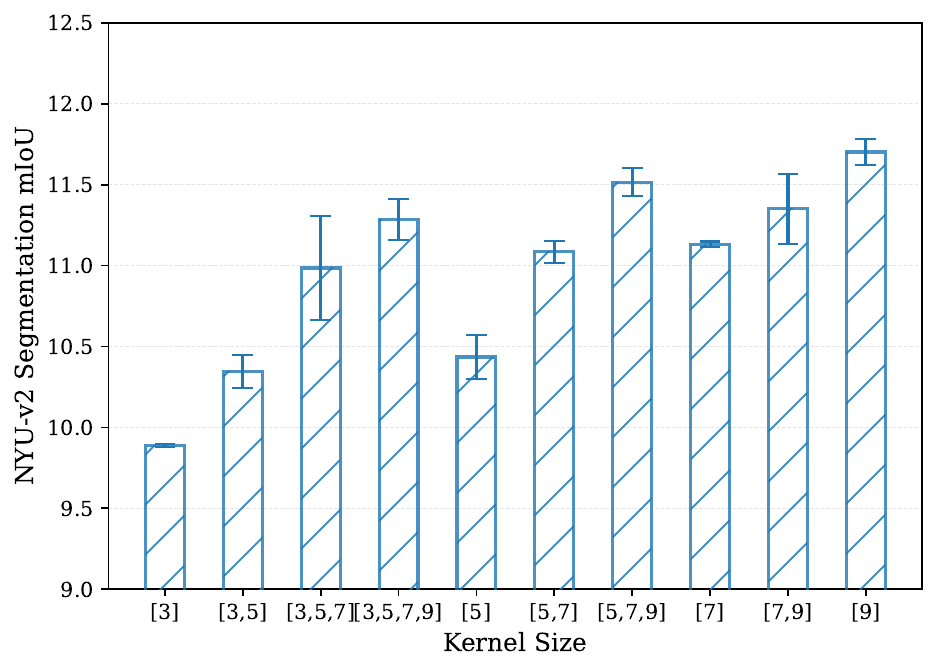} \hfill
        \includegraphics[width=.325\textwidth]{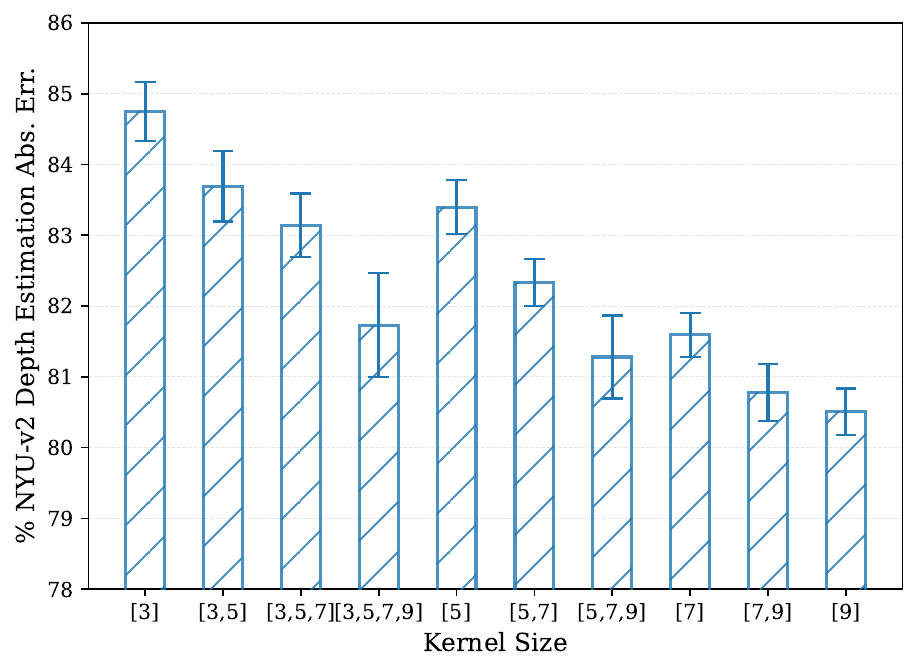} \hfill
        \includegraphics[width=.325\textwidth]{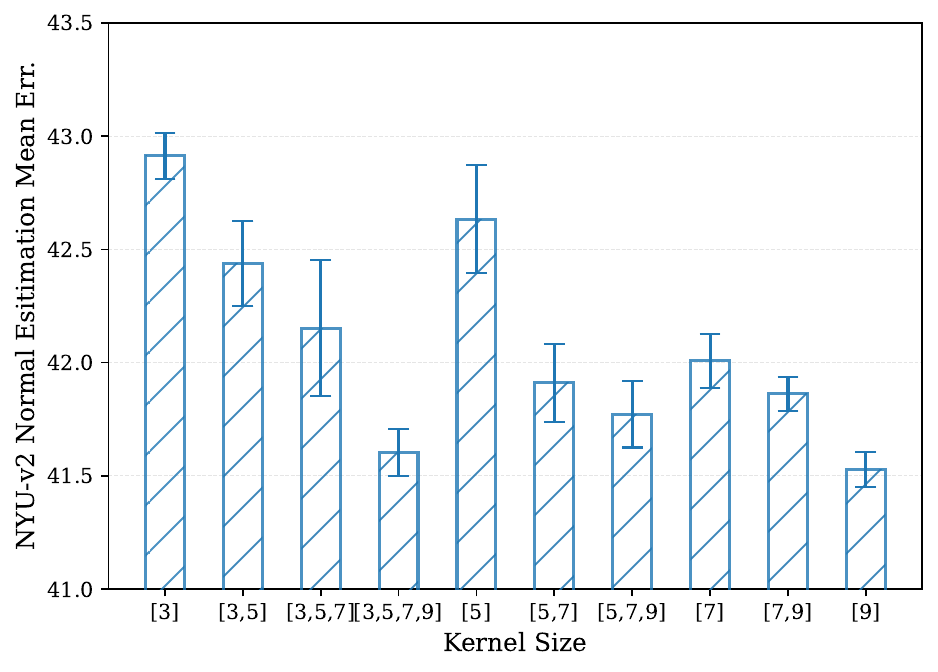}
        \caption{\scriptsize Avg. pooling \label{fig:dense_avg_pool}}
    \end{subfigure} \\
    \centering
    \begin{subfigure}[b]{0.485\textwidth}
        \centering
        \includegraphics[width=.325\textwidth]{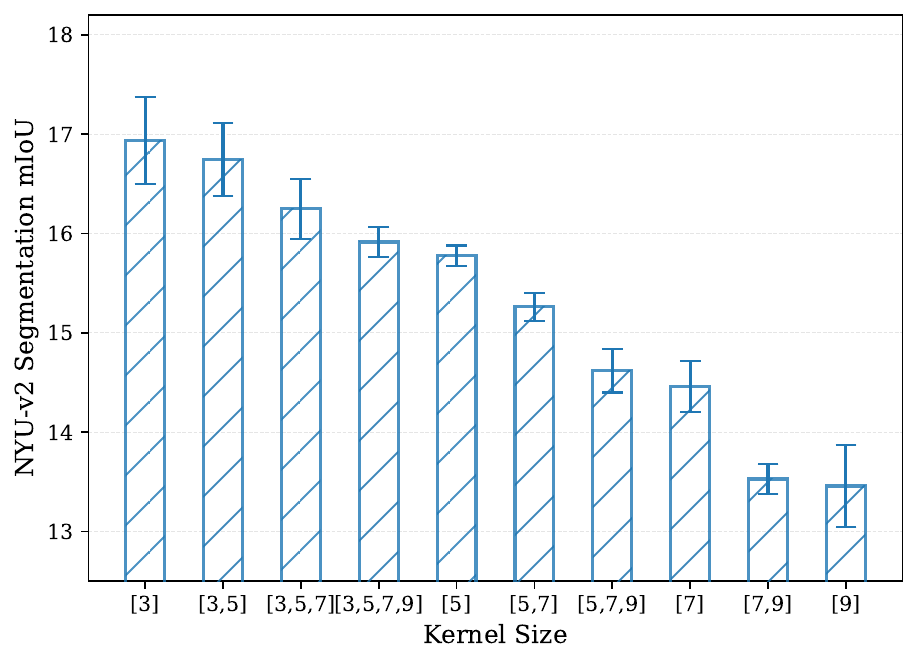} \hfill
        \includegraphics[width=.325\textwidth]{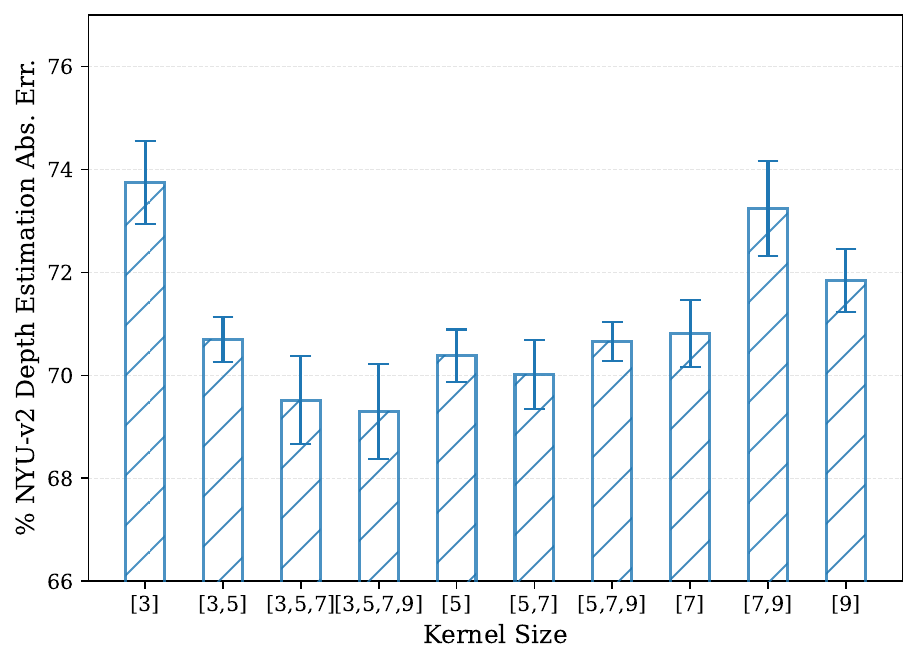} \hfill
        \includegraphics[width=.325\textwidth]{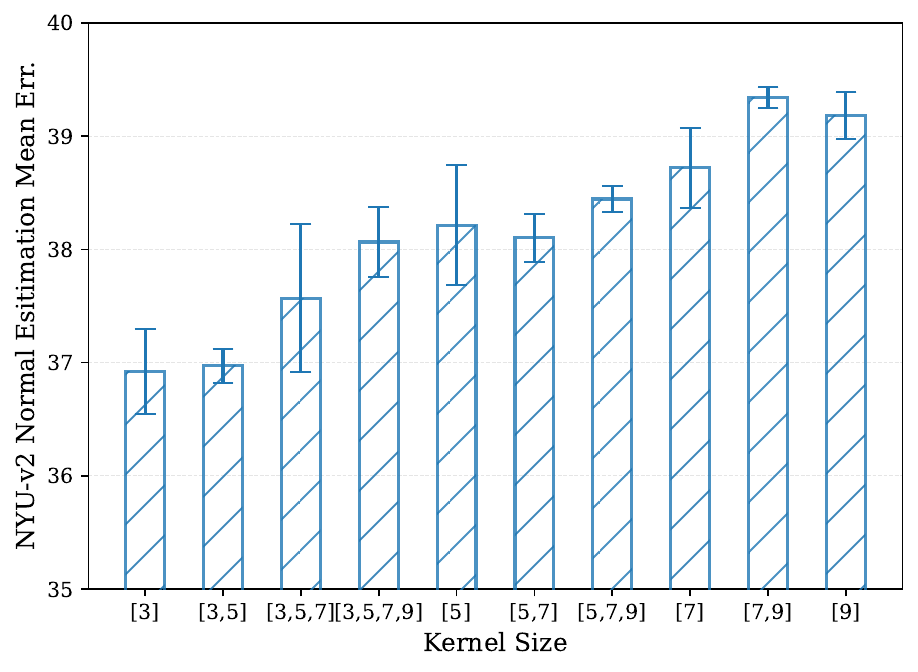}
        \caption{\scriptsize Max pooling \label{fig:dense_max_pool}}
    \end{subfigure} \\
    \centering
    \begin{subfigure}[b]{0.485\textwidth}
        \centering
        \includegraphics[width=.325\textwidth]{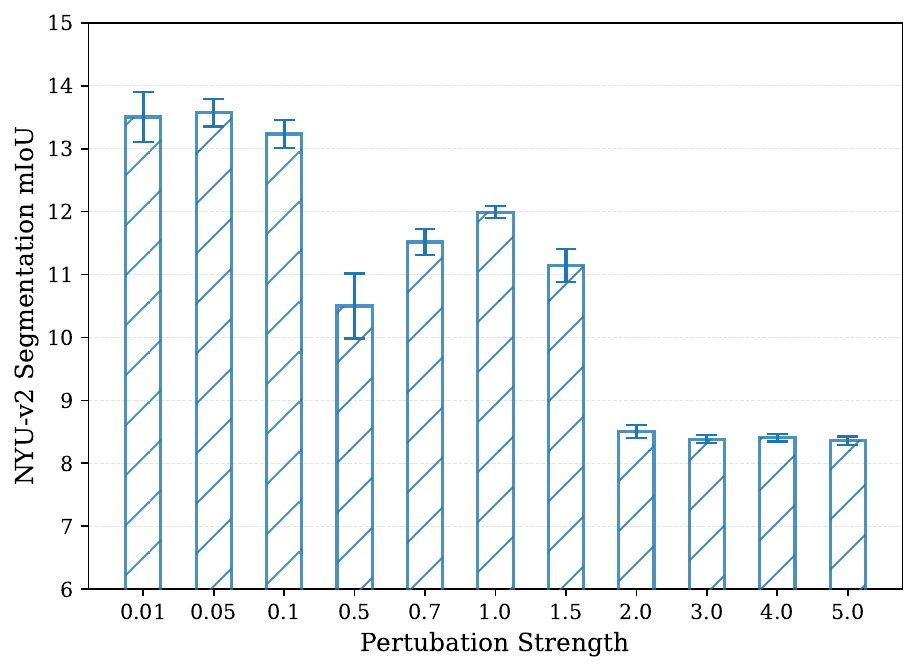} \hfill
        \includegraphics[width=.325\textwidth]{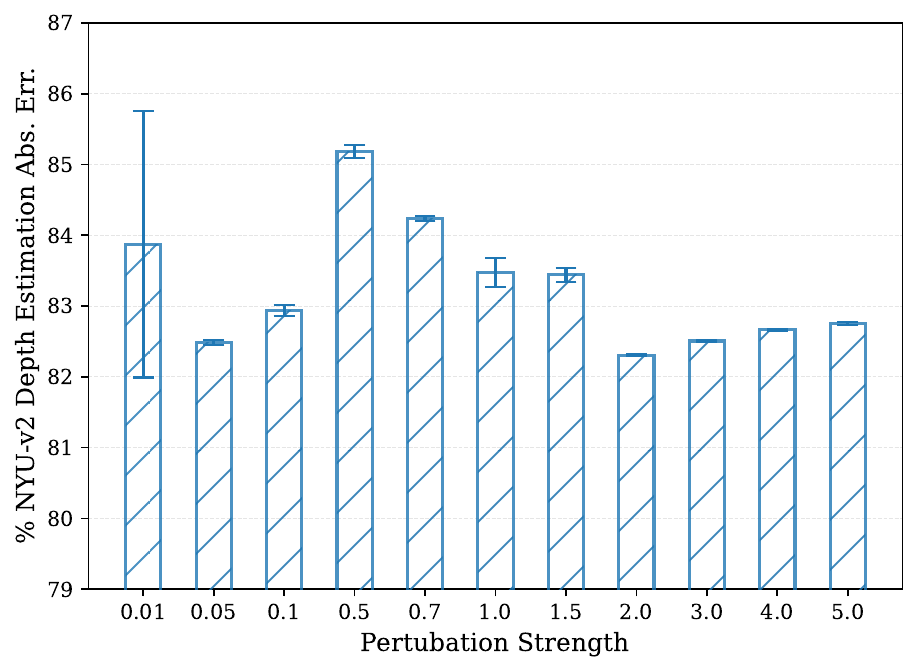} \hfill
        \includegraphics[width=.325\textwidth]{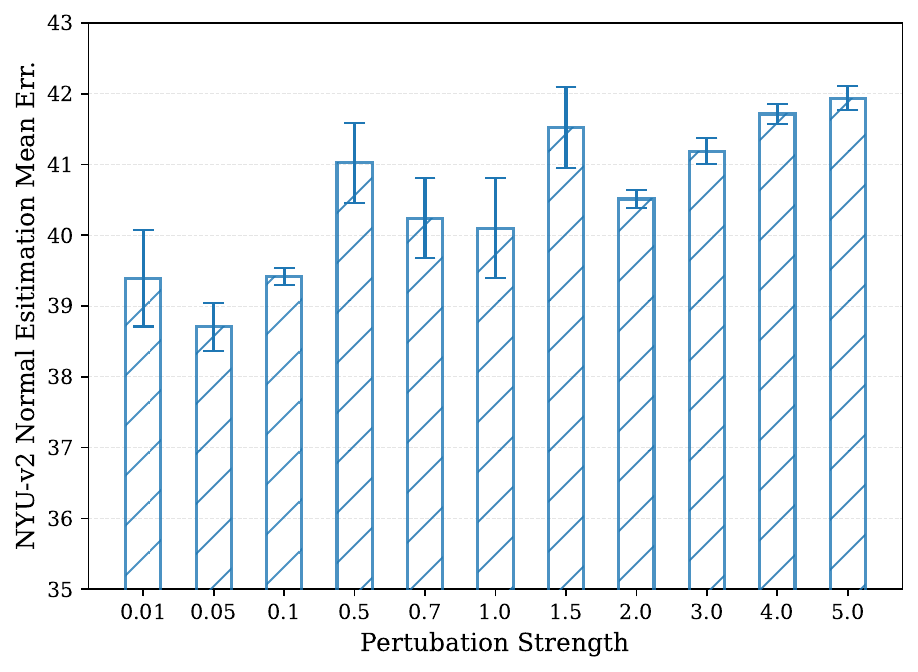}
        \caption{\scriptsize Perturbation \label{fig:dense_perturbation}}
    \end{subfigure} \\
    \centering
    \begin{subfigure}[b]{0.485\textwidth}
        \centering
        \includegraphics[width=.325\textwidth]{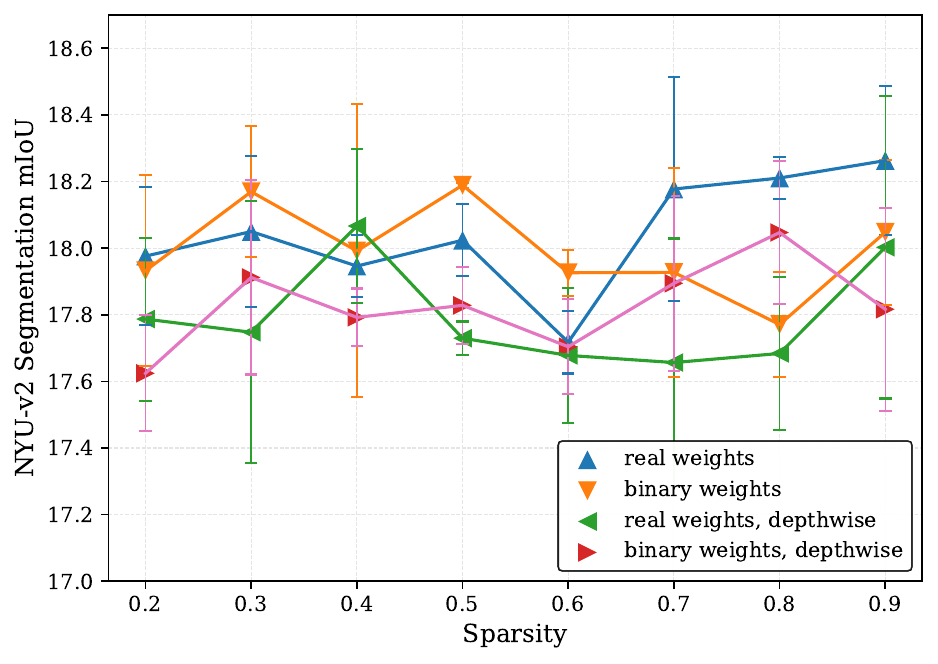} \hfill
        \includegraphics[width=.325\textwidth]{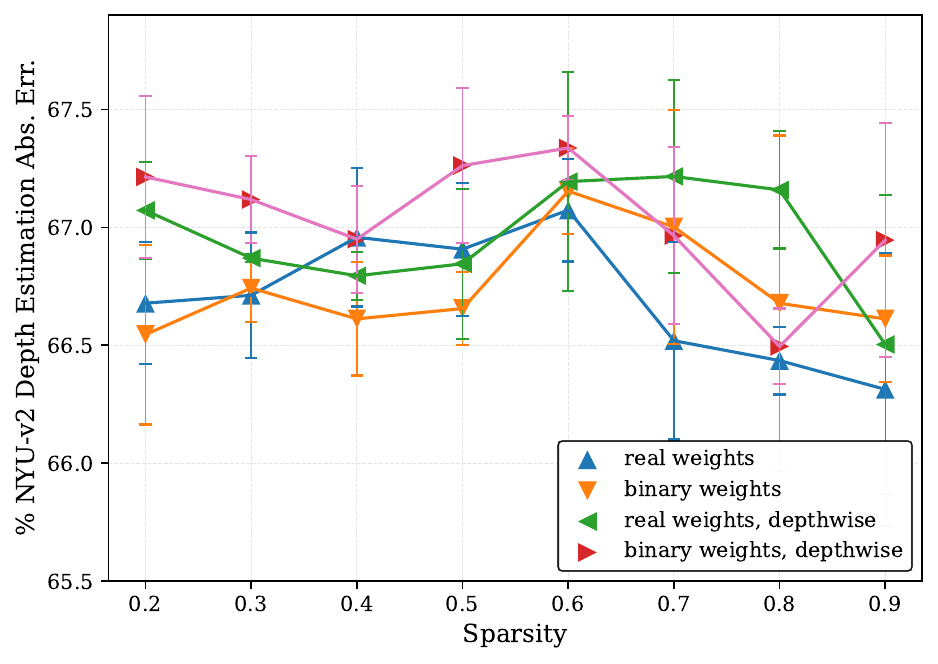} \hfill
        \includegraphics[width=.325\textwidth]{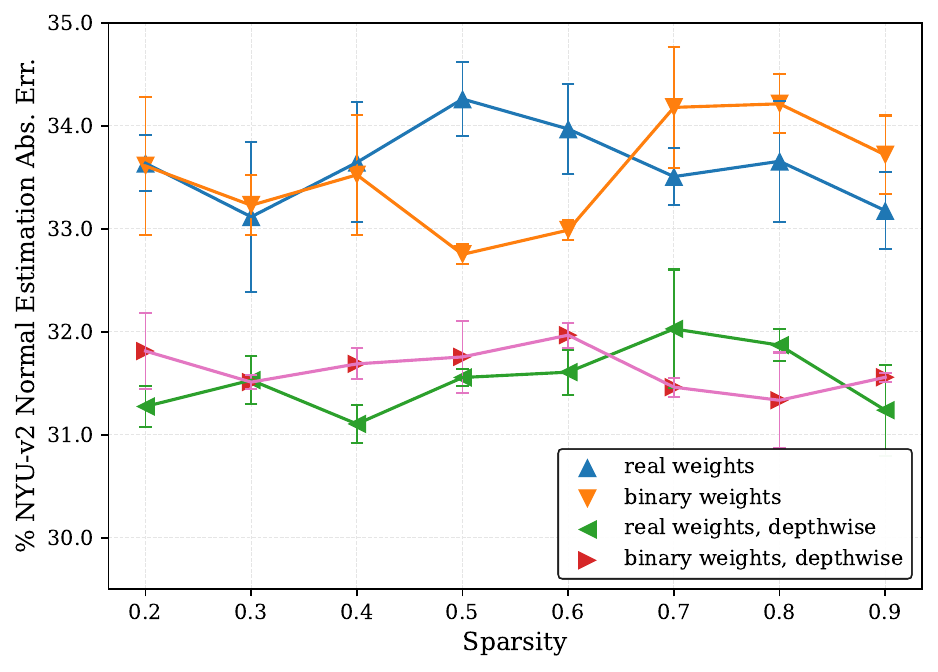}
        \caption{\scriptsize Convolution (sparsity) \label{fig:dense_conv}}
    \end{subfigure} \\
    \centering
    \begin{subfigure}[b]{0.485\textwidth}
        \centering
        \includegraphics[width=.325\textwidth]{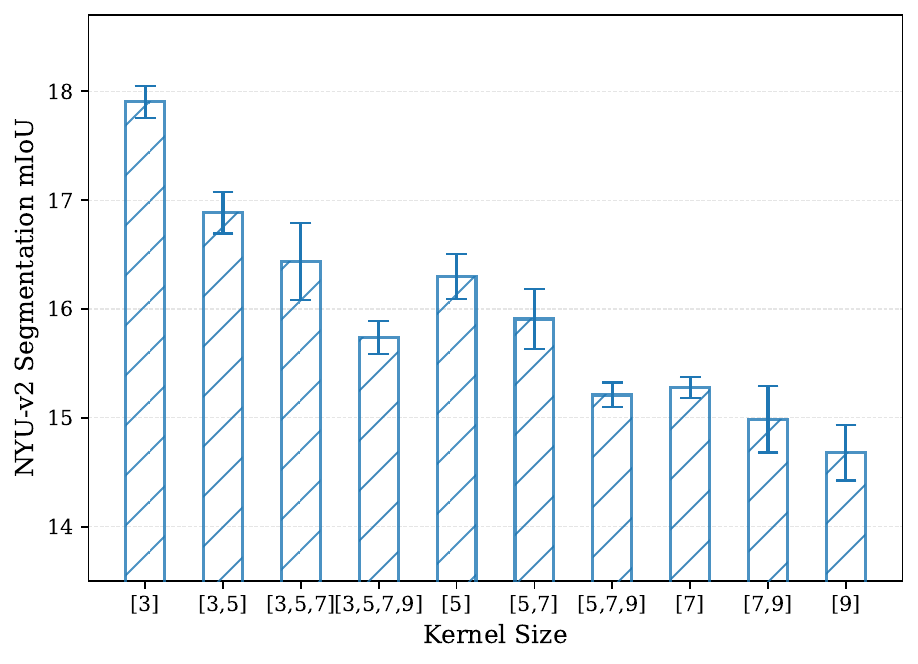} \hfill
        \includegraphics[width=.325\textwidth]{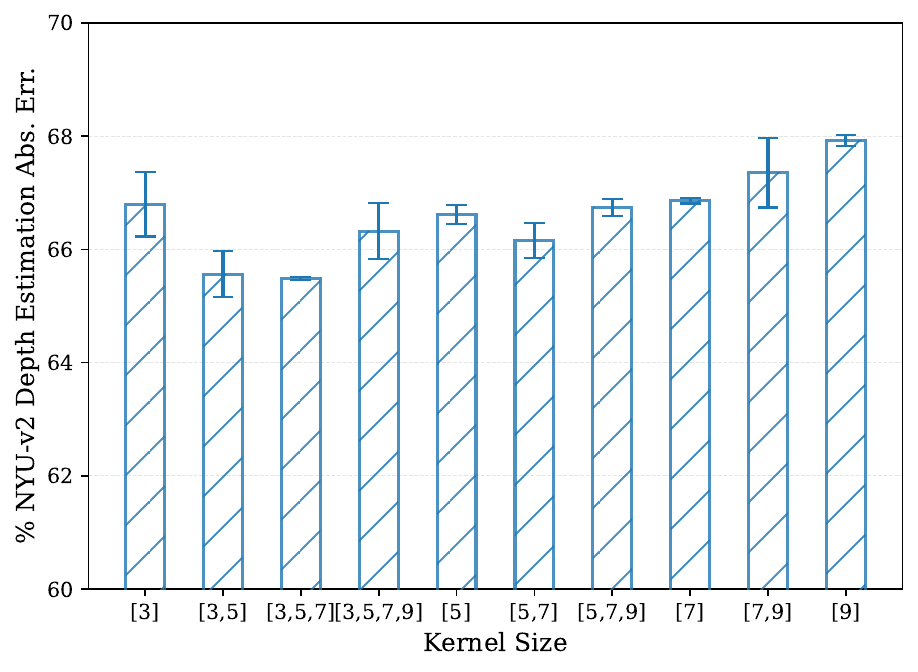} \hfill
        \includegraphics[width=.325\textwidth]{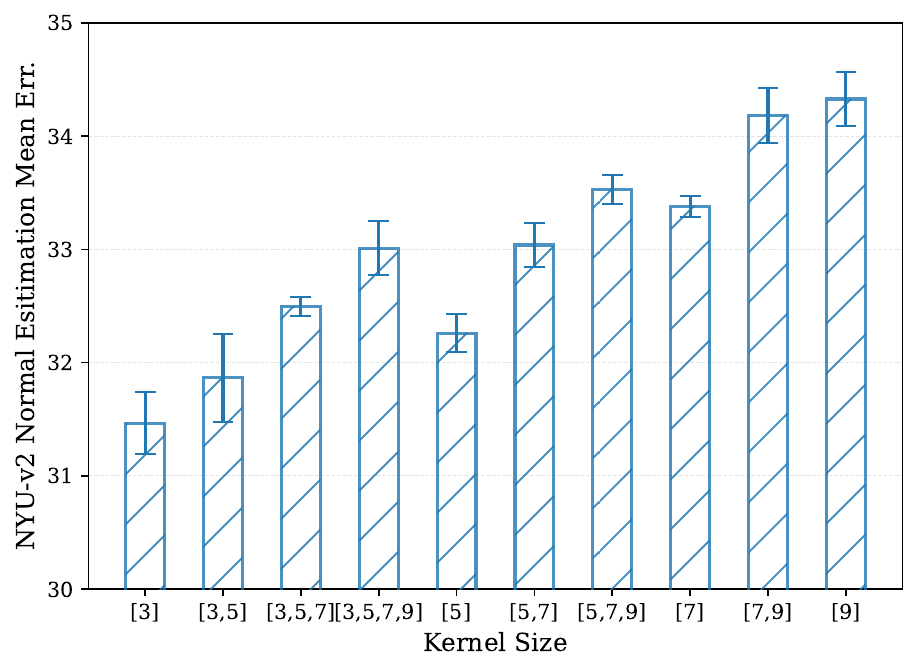}
        \caption{\scriptsize Convolution (kernel size) \label{fig:dense_conv_scale}}
    \end{subfigure}
    \caption*{Figure S1. Effect of different hyperparameters of individual NLP on NYU-v2 dense prediction MTL problem. For each sub-figure (a) - (e), we show the semantic segmentation mIoU ($\uparrow$), depth estimation absolute error ($\downarrow$), and surface normal estimation mean error ($\downarrow$).}
\end{figure}

\section{Description of Datasets and Baselines \label{sec:app_datasets_baselines}}
In this section, we first provide the additional details of CelebA and Cityscapes datasets in Table \ref{tab:app_celeba} and \ref{tab:app_cityscapes}, respectively. 
Then, we provide a brief overview of the baseline methods that we compared against in this work, as follows:

\begin{itemize}
    \item {STL}: single task learning with one network for each task.
    
    \item {Hard sharing}: standard multi-task learning, i.e., a fully shared network with uniform task weighting.
    
    \item {GradNorm} \cite{chen2018gradnorm}: a MTL method with a fully shared network and learnable tasks weighting.
    
    \item {MGDA-UB} \cite{MGDA}: a multi-objective alternative to MTL with a fully shared network.
    
    \item {Task Routing} \cite{Strezoski@Many}: a parameter partitioning method with randomly initialized binary masks.
    
    \item {Max. Roaming} \cite{pascal2020maximum}: another parameter partitioning method with dynamic masks.
    
    \item {Cross-stitch} \cite{Misra@Cross}: a soft-sharing method with feature fusion.
    \item {MTAN} \cite{liu2019end}: another soft-sharing method with attention.
    
\end{itemize}

\begin{table}[ht]
    \centering
    \caption{Details of the 40 attributes provided by the CelebA dataset \cite{liu@Deep}. For visualization purposes, we group them into eight categories. \label{tab:app_celeba}}
    \resizebox{.49\textwidth}{!}{%
    \begin{tabular}{@{\hspace{2mm}}c|l@{\hspace{2mm}}}
    \toprule
    \multicolumn{1}{c|}{\textbf{Group}} & \multicolumn{1}{c}{\textbf{40-attribute}} \\ \midrule
    global & \begin{tabular}[c]{@{}l@{}}attractive, blurry, chubby, double chin, heavy makeup, \\ male, oval face, pale skin, young\end{tabular} \\ \midrule
    eyes & \begin{tabular}[c]{@{}l@{}}bags under eyes, eyeglasses, narrow eyes, arched eyebrows,\\ bushy eyebrows\end{tabular} \\ \midrule
    hair & \begin{tabular}[c]{@{}l@{}}bald, bangs, black hair, blond hair, brown hair, gray hair, \\ receding hairline, straight hair, wavy hair\end{tabular} \\ \midrule
    mouth & big lips, mouth slightly open, smiling, wearing lipstick \\ \midrule
    nose & big nose, pointy nose \\ \midrule
    beard & 5 o'clock shadow, goatee, mustache, no beard, sideburns \\ \midrule
    cheek & high cheekbones, rosy cheeks \\ \midrule
    wearings & \begin{tabular}[c]{@{}l@{}}wearing earrings, wearing hat, wearing necklace, \\ wearing necktie\end{tabular} \\ \bottomrule
    \end{tabular}%
    }
\end{table}

\begin{table}[ht]
    \centering
    \caption{Three levels of semantic categories for the Cityscapes dataset \cite{Cordts@The,liu2019end}. The experimental results in this work are based on the 7-class setting. \label{tab:app_cityscapes}}
    \resizebox{.49\textwidth}{!}{%
    \begin{tabular}{@{\hspace{2mm}}l|l|l@{\hspace{2mm}}}
    \toprule
    \multicolumn{1}{c|}{\textbf{2-class}} & \multicolumn{1}{c|}{\textbf{7-class}} & \multicolumn{1}{c}{\textbf{19-class}} \\ \midrule
    \multirow{8}{*}{background} & void & void \\ \cmidrule(l){2-3} 
     & flat & road, sidewalk \\ \cmidrule(l){2-3} 
     & construction & building, wall, fence \\ \cmidrule(l){2-3} 
     & object & pole, traffic light, traffic sign \\ \cmidrule(l){2-3} 
     & nature & vegetation, terrain \\ \cmidrule(l){2-3} 
     & sky & sky \\ \midrule
    \multirow{2}{*}{foreground} & human & person, rider \\ \cmidrule(l){2-3} 
     & vehicle & carm truck, bus, caravan, trailer, train, motorcycle \\ \bottomrule
    \end{tabular}%
    }
\end{table}